\documentclass{article}

\usepackage{microtype}
\usepackage{graphicx}
\usepackage{subfigure}
\usepackage{booktabs} 
\usepackage{amssymb}
\usepackage{amsmath}
\usepackage{algorithm}
\usepackage{algorithmic}
\usepackage{adjustbox}
\usepackage{multirow}
\usepackage{enumerate}
\usepackage{pifont}
\newcommand{\cmark}{\ding{51}}%
\newcommand{\xmark}{\ding{55}}%

\usepackage{hyperref}

\usepackage[accepted]{icml2023}

\usepackage{amsmath}
\usepackage{amssymb}
\usepackage{mathtools}
\usepackage{amsthm}

\usepackage[capitalize,noabbrev]{cleveref}

\theoremstyle{plain}

\theoremstyle{definition}

\theoremstyle{remark}

\usepackage[textsize=tiny]{todonotes}

\icmltitlerunning{Reconstructive Neuron Pruning for Backdoor Defense}

\begin{document}

\twocolumn[
\icmltitle{Reconstructive Neuron Pruning for Backdoor Defense}



\icmlsetsymbol{equal}{*}

\begin{icmlauthorlist}
\icmlauthor{Yige Li}{yyy}
\icmlauthor{Xixiang Lyu}{yyy}
\icmlauthor{Xingjun Ma}{comp,ailab}
\icmlauthor{Nodens Koren}{sch}
\icmlauthor{Lingjuan Lyu}{aaa}
\icmlauthor{Bo Li}{zzz}
\icmlauthor{Yu-Gang Jiang}{comp,ailab}
\end{icmlauthorlist}

\icmlaffiliation{yyy}{Xidian University}
\icmlaffiliation{comp}{Fudan University}
\icmlaffiliation{ailab}{Shanghai Artificial Intelligence Laboratory}
\icmlaffiliation{sch}{University of Copenhagen}
\icmlaffiliation{aaa}{Sony AI}
\icmlaffiliation{zzz}{University of Illinois at Urbana–Champaign}

\icmlcorrespondingauthor{Xixiang Lyu}{xxlv@mail.xidian.edu.cn}
\icmlcorrespondingauthor{Xingjun Ma}{xingjunma@fudan.edu.cn}

\icmlkeywords{Machine Learning, ICML}

\vskip 0.3in
]



\printAffiliationsAndNotice{} 

\begin{abstract}
Deep neural networks (DNNs) have been found to be vulnerable to backdoor attacks, raising security concerns about their deployment in mission-critical applications. While existing defense methods have demonstrated promising results, it is still not clear how to effectively remove backdoor-associated neurons in backdoored DNNs.
In this paper, we propose a novel defense called \emph{Reconstructive Neuron Pruning} (RNP) to expose and prune backdoor neurons via an unlearning and then recovering process. 
Specifically, RNP first unlearns the neurons by maximizing the model's error on a small subset of clean samples and then recovers the neurons by minimizing the model's error on the same data. In RNP, unlearning is operated at the neuron level while recovering is operated at the filter level, forming an asymmetric reconstructive learning procedure.
We show that such an asymmetric process on only a few clean samples can effectively expose and prune the backdoor neurons implanted by a wide range of attacks, achieving a new state-of-the-art defense performance. Moreover, the unlearned model at the intermediate step of our RNP can be directly used to improve other backdoor defense tasks including backdoor removal, trigger recovery, backdoor label detection, and backdoor sample detection.
Code is available at \url{https://github.com/bboylyg/RNP}.
\end{abstract}

\section{Introduction}
Over the past decade, deep neural networks (DNNs) have demonstrated groundbreaking performance in solving various complex real-world problems. However, despite these significant successes, recent works have shown that DNNs are susceptible to different types of adversaries \cite{szegedy2013intriguing,gu2017badnets}. Backdoor attacks represent one such type of adversary that injects malicious triggers into victim models during training either by poisoning the training data or manipulating the training procedure with specifically-designed trigger patterns. A backdoored model functions normally with clean data but predicts the pre-specified backdoor class whenever the trigger patterns appear. In this era, where pre-trained models and outsourced training via Machine Learning as a Service (MLaaS) are commonly adopted to achieve optimal performance at a minimal cost, the threats posed by backdoor attacks have increasingly become an issue we cannot overlook.

Numerous methods have been proposed to defend against backdoor attacks, with detection and removal methods being the two primary types. Detection methods identify whether a model has been backdoored \cite{wang2019neural,guo2019tabor,liu2019abs,xu2021detecting,shen2021backdoor,hu2022trigger} or whether a test sample contains a backdoor trigger \cite{tran2018spectral,chen2018detecting,tang2021demon,zeng2021rethinking,chen2022effective}. Removal (or mitigation) methods are effective in purifying backdoored models by balancing the backdoor effect and their clean performance \cite{liu2018fine,li2021neural,wu2021adversarial}. It is often observed in these methods that a backdoored DNN contains both clean and backdoor neurons, with the backdoor neurons being activated only by trigger patterns. The study conducted in \cite{wu2021adversarial} also shows that backdoor neurons are more sensitive to adversarial perturbations. Intuitively, if backdoor neurons can be accurately identified from a backdoored model, we can immediately prune those neurons to obtain a clean model. 

To this end, we propose a novel method called \emph{Reconstructive Neuron Pruning} (RNP) to expose and prune backdoor neurons from a backdoored model by unlearning and then recovering the neurons. Specifically, given a backdoored model, RNP first unlearns the model by maximizing its error on clean samples through gradient ascent and then recovers (relearns) the neurons by minimizing the model's error on the same samples. Interestingly, we find that if the unlearning is performed at the neuron level while the recovering is performed at the filter level, then the network tends to relocate the backdoor neurons to compensate for the loss of clean features caused by the unlearning. Such an asymmetric operation can be very effective in locating the backdoor neurons with only a few clean samples (e.g., 500 images for the CIFAR-10 dataset). The backdoor neurons can then be easily pruned from the network.

With RNP, we have conducted so far the most extensive defense experiments in the current literature against 12 advanced backdoor attacks. Empirical results across different datasets and model architectures show that our RNP outperforms the current state-of-the-art method ANP~\cite{wu2021adversarial} against 9/12 attacks on CIFAR-10 dataset and 5/5 attacks on an ImageNet subset. In some cases, our RNP only needs to remove 41 neurons from the backdoored (by BadNets~\cite{gu2017badnets}) model to reduce the attack success rate from 100\% to 0.20\%, while causing no significant clean accuracy degradation. Moreover, we demonstrate that the unlearned model obtained at the intermediate step of RNP can be directly used to improve other backdoor defense tasks.

To summarize, our main contributions are: 

\begin{itemize}
\item We introduce the novel technique of neuron unlearning and recovering on the same set of samples and reveal that such a simple reconstruction-based learning process can help expose backdoor neurons in DNNs.

\item We propose a new defense method called \emph{Reconstructive Neuron Pruning} (RNP), which detects and prunes backdoor neurons via a neuron-level unlearning followed by a filter-level recovering with the help of a few clean samples.

\item We empirically show that RNP outperforms existing backdoor defenses by a considerable margin against 12 advanced backdoor attacks, and that the unlearned models can aid in trigger recovery, backdoor label detection, and backdoor sample detection.

\end{itemize}

\section{Related Work}

\subsection{Backdoor Attack}
 Depending on how the trigger pattern is crafted, existing backdoor attacks can be primarily categorized into two types: input-space attacks and feature-space attacks.

\textbf{Input-space attacks.} This type of attack injects a pre-defined trigger pattern into a small proportion of the training data to trick the model into learning the correlation between the trigger pattern and the backdoor label. The trigger pattern can be relatively simple, such as a single pixel \cite{tran2018spectral}, a black-white square \cite{gu2017badnets}, random noise \cite{chen2017targeted}, or more complex patterns such as adversarial perturbation \cite{turner2019clean}, natural reflection \cite{liu2020reflection}, and sample-wise patterns \cite{nguyen2020input,li2021invisible, wang2022invisible}.

\textbf{Feature-space attacks.} These attacks directly manipulate the training procedure to optimize the backdoor objective in the feature space \cite{shafahi2018poison, cheng2021deep, zhao2022defeat} or directly modify the model parameters via weight perturbation \cite{garg2020can, qi2022towards}. These two types of attacks represent two typical threat models: input-space attacks only need access to a small subset of the training data, while feature-space attacks require full access to the training procedure or the final model. Input-space attacks could occur during the data collection process, while feature-space attacks could occur in outsourced training via MLaaS or when downloading pre-trained models from untrusted sources. As we will show in our experiments, feature-space attacks are generally more difficult to defend against than input-space attacks.

\begin{table}[!tp]
\caption{Whether a defense technique can help backdoor detection (BD), trigger recovery (TR), or backdoor removal (BR). }
\centering
\small
\begin{sc}
    \begin{tabular}{cccc}
    \toprule
    Defense & BD & TR & BR \\ \midrule
    NC & \cmark & \cmark & \xmark \\
    STRIP & \cmark & \xmark & \xmark \\
    Fine-pruning & \cmark & \xmark & \xmark \\
    ABL & \xmark & \xmark & \cmark \\
    I-BAU & \xmark & \xmark & \cmark \\
    ANP & \xmark & \xmark & \cmark \\
    RNP (Ours) & \cmark & \cmark & \cmark \\ \bottomrule
    \end{tabular}
\end{sc}
\vskip -0.1in
\label{tab:functions}
\end{table}

\begin{figure*}[!tp]
\small
\centering
\centerline{\includegraphics[width = .95\linewidth]{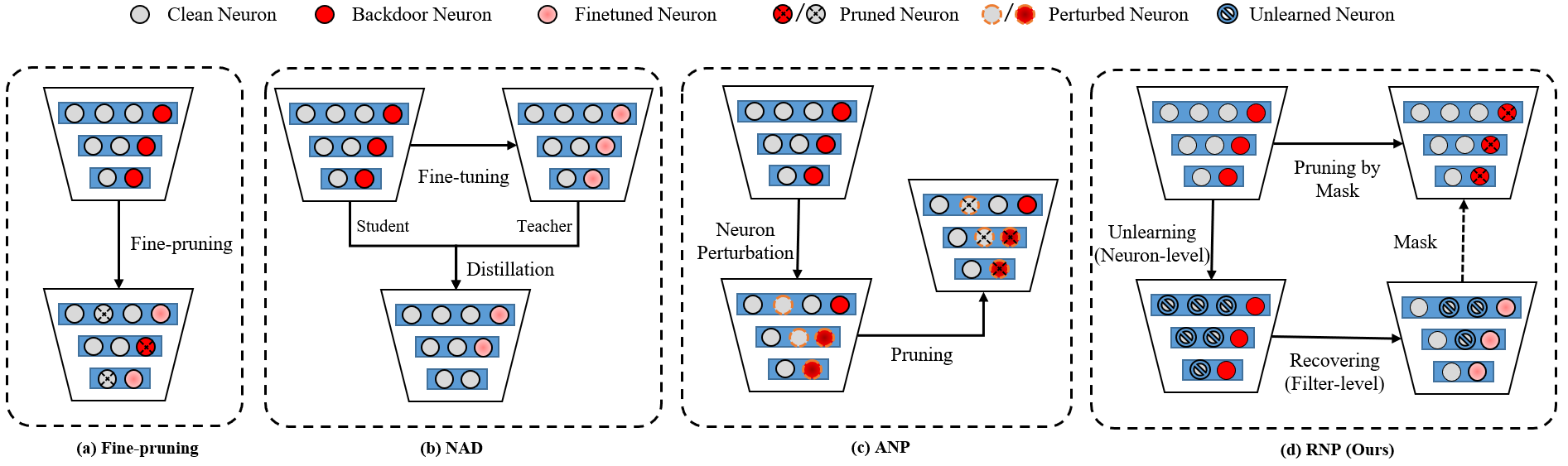}}
\caption{Overview of our proposed \textbf{RNP} framework, in comparison with 3 existing backdoor removal methods: \textbf{Fine-pruning}, \textbf{NAD}, and \textbf{ANP}. RNP exposes the backdoor via a neuron-level unlearning followed by a filter-level recovering. Pruning is then applied to remove the exposed backdoor neurons. Note that both ANP and our RNP do not need fine-tuning after the pruning.}
\label{fig:overview}
\end{figure*}

\subsection{Backdoor Defense} 
Numerous approaches have been proposed to defend DNNs against backdoor attacks, among which backdoor detection and removal methods are the two most popular ones. 

\textbf{Backdoor detection.} 
Several detection works identify backdoors based on the prediction bias on different input examples \cite{li2020rethinking} or the statistical deviation in the feature space \cite{tran2018spectral, chen2018detecting, liu2022complex}.
More effective detection methods leverage reverse engineering techniques to recover the trigger pattern and then identify the backdoor label by anomaly detection \cite{wang2019neural,liu2019abs,guo2019tabor,hu2022trigger}. The most representative method is Neural Cleanse (NC) \cite{wang2019neural}, which recovers trigger patterns that can alter the model's predictions with minimum perturbation. There are also methods that detect backdoored samples at inference time, for example, the STRIP method \cite{gao2019strip}.

\textbf{Backdoor removal.} Backdoor removal methods aim to erase the backdoors from backdoored models without significantly reducing their performance on clean samples. This line of work includes Fine-tuning, Fine-pruning \cite{liu2018fine}, Neural Attention Distillation (NAD) \cite{li2021neural}, Channel Lipschitzness Pruning \cite{zheng2022data}, and Shapley Estimation \cite{guan2022few}. More recently, training-time defense methods \cite{li2021anti, huang2022backdoor, wang2022training} have been proposed to train clean models directly on backdoored data. Meanwhile, Adversarial Unlearning of Backdoors via Implicit Hypergradient (I-BAU) \cite{zeng2021adversarial} is proposed to cleanse backdoored models with adversarial training. Another recent work, Adversarial Neuron Pruning (ANP) \cite{wu2021adversarial}, prunes those neurons that are more adversarially sensitive to remove backdoors. ANP has achieved a new state-of-the-art result against input-space attacks. Our experiments will show that ANP fails on more advanced feature-space attacks, indicating that adversarial perturbation may not be an ideal approach for exposing backdoor neurons that are deeply intertwined with the features. Table \ref{tab:functions} summarizes the benefits of existing and our proposed RNP defense methods.

\section{Proposed Method}
In this section, we first introduce the threat model, then present our proposed defense, \emph{Reconstructive Neuron Pruning} (RNP), and an illustrative example.

\subsection{Threat Model} 
We assume the adversary has successfully injected a backdoor trigger into the target model. The defender's goal is to remove the backdoor trigger from the target model with minimal impact on its clean accuracy (accuracy on clean test samples), and at the same time, reveal as much information about the attack as possible, such as the trigger pattern and the backdoor label. Following prior works \cite{li2021neural,wu2021adversarial}, we assume the defender has a small subset of clean data (e.g., 1\% of carefully-examined training data) to develop the defense strategy, which we call \emph{defense data}.

\subsection{Reconstructive Neuron Pruning}
\noindent\textbf{Overview.} 
Figure \ref{fig:overview} illustrates our proposed RNP defense. At the core of RNP is a reconstructive learning process that first unlearns the neurons on the defense data via \textit{Neuron Unlearning} (NU) and then recovers the neurons on the same data via \textit{Filter Recovering} (FR).
As the defense data is clean, NU tends to unlearn primarily the clean neurons, i.e., neurons associated with the clean features. The backdoor neurons, i.e., neurons associated with the backdoor features, are largely preserved in the unlearned model. As such, the unlearned model can be leveraged to improve other analyses such as trigger recovery, backdoor label detection, and backdoor sample detection.

The mechanisms of existing methods, Fine-pruning, NAD, and ANP are also illustrated in Figure \ref{fig:overview} for comparison. Fine-pruning is a conventional pruning method that prunes those small-norm neurons from the backdoored model, while NAD \cite{li2021neural} adopts the fine-tuned model as a teacher to distill neurons of the backdoored model. ANP \cite{wu2021adversarial} exploits adversarial perturbations to find neurons that are more sensitive to adversarial perturbations as backdoor neurons. Compared to the adversarial perturbation technique used by ANP, our RNP, with the asymmetric unlearning and recovering procedure, exposes more backdoor-associated neurons and leads to better backdoor purification at the pruning step.

\noindent\textbf{Unlearning.}
Consider a standard $K$-class classification task on poisoned training data $\mathcal{D} = \mathcal{D}_{c} \cup \mathcal{D}_{b}$, with $\mathcal{D}_{c}=\{(\boldsymbol{x}_c^{(i)}, y_{c}^{(i)})\}_{i=1}^{N}$ representing the clean subset and  $\mathcal{D}_{b}=\{(\boldsymbol{x}_b^{(j)}, y_{b}^{(j)})\}_{j=1}^{M}$ representing the backdoor subset. 
A backdoor attack, or training a backdoored model, can be viewed as a dual-task learning problem defined on both the clean subset $\mathcal{D}{c}$ (clean task) and the backdoor subset $\mathcal{D}{b}$ (backdoor task) as follows:

\begin{equation} \label{eq:1}
\begin{aligned}
\underset{\theta=\theta_{c}\cup\theta_{b}}{\arg \min } & \Big[ \underbrace{ \mathbb{E}_{(\boldsymbol{x}_c, y_c)\in \mathcal{D}_{c}}\mathcal{L}(f(\boldsymbol{x}_c,\ y_{c};\ \theta_{c}))}_{\text{clean task}} \\
& + \underbrace{ \mathbb{E}_{(\boldsymbol{x}_b, y_b)\in \mathcal{D}_{b}} \mathcal{L}(f(\boldsymbol{x}_b,\ y_{b};\ \theta_{b})) }_{\text{backdoor task}} \Big],
\end{aligned}
\end{equation}
where $f$ is the victim model with parameters $\theta$, $\mathcal{L}$ denotes the classification loss (e.g., cross-entropy), and $\theta_{c}$ and $\theta_{b}$ denote the clean and the backdoor neurons, respectively. The parameter space of the backdoored model can be decomposed into $\theta={\theta_{c} \cup \theta_{b}}$ because there exists a high level of independence between the clean task and the backdoor task \cite{li2021neural}, i.e., backdoor attacks by design should not impact the model's performance on clean samples \cite{gu2017badnets}. Note that, although $\theta=\theta_{c}\cup\theta_{b}$, it does not mean $\theta_{c}$ cannot overlap with $\theta_{b}$, i.e., it is possible that $\theta_{c}\cap\theta_{b}\neq \emptyset$.

Intuitively, a model can be unlearned with respect to a certain task by maximizing its loss on the data that defines the task, an opposite process to model training.
Following the above formulation, there exist two possible strategies to unlearn a backdoored model: 1) maximize the model's loss on the backdoor data $\mathcal{D}_{b}$, or 2) maximize the model's loss on the defense data $\mathcal{D}_{d}$. The first strategy is infeasible for backdoor defense as the defender does not know the backdoor data in advance. This naturally leads us to the second strategy that solves the following maximization problem:

\begin{equation} \label{eq:2}
\begin{aligned}
\underset{\mathbf{\theta}}{\max } \; \mathbb{E}_{(\boldsymbol{x}_d,\ y_{d}) \in \mathcal{D}_{d}} \mathcal{L} (f\left(\boldsymbol{x}_d,\ y_{d} ; \theta \right)),
\end{aligned}
\end{equation}
where $\mathcal{L}$ is the cross-entropy loss, and $(\boldsymbol{x}_d,\ y_{d})$ are the clean samples in the defense dataset and their labels. We denote the parameters of the unlearned model by $\hat{\theta}$. Note that the unlearning is performed at a neuron level for all neurons of the model, so it is termed as \textit{Neuron Unlearning (NU)}.

The above unlearning strategy is extremely simple but can be very effective. In fact, it is not surprising that clean neurons will be effectively unlearned if the above maximization is applied to the entire clean data $\mathcal{D}_{c}$. Interestingly, we find that the unlearning can be easily achieved with only a few clean samples, i.e., the defense data required by the defender is extremely small (i.e., $|\mathcal{D}_{d}| \ll |\mathcal{D}|$). For example, on the CIFAR-10 dataset, 1\% of the clean training data is sufficient to expose the backdoor neurons injected by a wide range of advanced backdoor attacks. Moreover, the unlearning can be safely terminated when the performance of the model on defense data $\mathcal{D}_{d}$ is close to random guess. Note that the backdoor label inference step is performed on the unlearned model with the defense data, i.e., the unlearned model tends to predict the backdoor label for all defense samples (one unique property of the unlearned model as further explained in Appendix \ref{ap:NU_label}).

\begin{algorithm} [!tp]
	\renewcommand{\algorithmicrequire}{\textbf{Input:}}
	\renewcommand{\algorithmicensure}{\textbf{Output:}}
	\caption{Reconstructive Neuron Pruning (RNP)}
	\begin{algorithmic}[1]
        \REQUIRE A backdoored model $f_{\theta}(\cdot)$ with parameter $\theta$, the total number of classes $K$, defense data $\mathcal{D}_{d}$, learning rate $\eta$, clean accuracy threshold $CA_{min}$, dynamic threshold $DT$ in [0, 1]
 
		\STATE Sample a mini-batch $(X_{d}, Y_{d})$ from $\mathcal{D}_{d}$
		
	\textbf{\# Neuron-level unlearning} 
		\REPEAT
        \STATE $\hat{\theta} \gets \underset{\mathbf{\theta}}{\max } \; \mathcal{L} (f\left(\boldsymbol{x}_d,\ y_{d} ; \theta \right))$
        \UNTIL $f_{\hat{\theta}}$'s clean accuracy $CA_{f_{\hat{\theta}}}(\mathcal{D}_{d}) \leq CA_{min}$
        \STATE Backdoor label: $y_{t} = \underset{K}{\arg \max} \ f(\boldsymbol{x}_d;\ \hat{\theta})$
	
	\textbf{\# Filter-level recovering} 
		\STATE $\mathbf{m}^{\kappa} = [1]^n$  \; \# initialized to be all ones
		\REPEAT
        \STATE $\mathbf{m}^{\kappa} = \mathbf{m}^{\kappa} - \eta \frac{\partial \mathcal{L}(f(X_{d}, Y_{d};\ \mathbf{m}^{\kappa} \odot \hat{\theta}))}{\partial \mathbf{m}^{\kappa}}$
        \STATE $\mathbf{m}^{\kappa} = clip_{[0,1]}(\mathbf{m}^{\kappa})$  \; \# 0-1 clipping
        \UNTIL training converged
	
	\textbf{\# Pruning}
	    \STATE $\mathbf{m}^{\kappa}=\mathbb{I}\left(\mathbf{m}^{\kappa} > DT \right)$  \# binarization for pruning
       \ENSURE $f_{\mathbf{m}^{\kappa} \odot \theta}$, $y_{t}$
	\end{algorithmic}
		\label{alg:1}
\end{algorithm}

\noindent\textbf{Recovering.}
This step aims to recover the clean features (features of the clean samples) erased by the previous unlearning step. Recovering can be effectively done by updating the unlearned model to minimize its classification loss on the defense data, a process similar to model training. In our RNP, the recovery is performed on the filters rather than the neurons, and a filter mask is used to locate potential backdoor filters (and their associated neurons). We term this technique as \emph{Filter Recovering (FR)}. Formally, FR solves the following minimization problem to learn the mask:

\begin{equation}\label{eq:3}
\begin{aligned}
\min_{\mathbf{m}^{\kappa} \in[0,1]^{n}} \mathbb{E}_{(\boldsymbol{x}_d, / y_d) \in \mathcal{D}_d}\mathcal{L}(f(\boldsymbol{x}_d,\ y_{d} ;\mathbf{m}^{\kappa} \odot \hat{\theta})),
\end{aligned}
\end{equation}
where $\boldsymbol{x}_d \in \mathcal{D}_{d}$ is the defense data, $\mathcal{L}$ is the cross-entropy loss, $\hat{\theta}$ are the parameters of the unlearned model obtained via NU, and $\mathbf{m}^{\kappa}$ is a mask applied on the filters. Finding the optimal mask $\mathbf{m}^{\kappa}$ can be viewed as a process to restore as many clean filters as possible to recover the model's clean features on $\mathcal{D}_{d}$.

The question is why recovering should be applied to the filters. Intuitively, if unlearning is also applied to the neurons, it will become a direct reversal of the unlearning, ending up with the unlearned clean neurons restored to their original values. This does not help expose the backdoor neurons. Filter-level recovering restricts the freedom of recovery to a coarser granularity, thus forcing the model to reuse the dormant backdoor neurons to compensate for the loss of the clean features. The clean neurons in this process are also relearned to recover the clean functionality, which, however, is not sufficient due to the limitation of the filter-level recovery. Neuron unlearning and filter recovering form an asymmetric unlearning-recovering (maximization-minimization) mechanism to expose backdoor neurons. We will provide empirical understandings of the mechanism and the necessity of unlearning-recovering in Section \ref{sec:understanding}.

\noindent\textbf{Pruning.}
The elements in the filter mask $\mathbf{m}^{\kappa}$ are all initialized to be one and clipped to be within the range of $[0, 1]$ during the recovering process. After recovering, a low value close to zero in $\mathbf{m}^{\kappa}$ indicates that the filter (and its associated neurons) contains mostly repurposed neurons that are likely to be backdoor-related. These neurons can thus be pruned to purify the backdoored model. As shown in Figure \ref{fig:motivation}, in the recovered model, the activations related to the trigger pattern are greatly decreased (mask value decreases to almost zero) while those of the clean features are significantly boosted (mask value remains close to 1 due to the clipping operation).

The complete procedure of our RNP is described in Algorithm \ref{alg:1}, where pruning is applied to the original model $f_{\theta}$ based on the learned filter mask $\mathbf{m}^{\kappa}$. The optimal pruning rate can be flexibly determined via dynamic thresholding in $[0, 1]$. The idea is to prune as many neurons as possible until the drop in the clean accuracy becomes unacceptable. In our experiments, we take dynamic thresholding as our default setting, unless otherwise stated. More analyses of different pruning rate determination strategies and the number of neurons pruned by our RNP can be found in Table \ref{tab:pruning_type} (see Appendix \ref{ap:pruning_type}) and Table \ref{tab:pruning_number}, respectively.

\subsection{An Illustrative Example} \label{sec:motivating}
Here, we provide an illustrative example to understand and verify that asymmetric unlearning and recovering can expose backdoor neurons.

\begin{figure}[!tp]
\small
\centering
\centerline{\includegraphics[width = .99\linewidth]{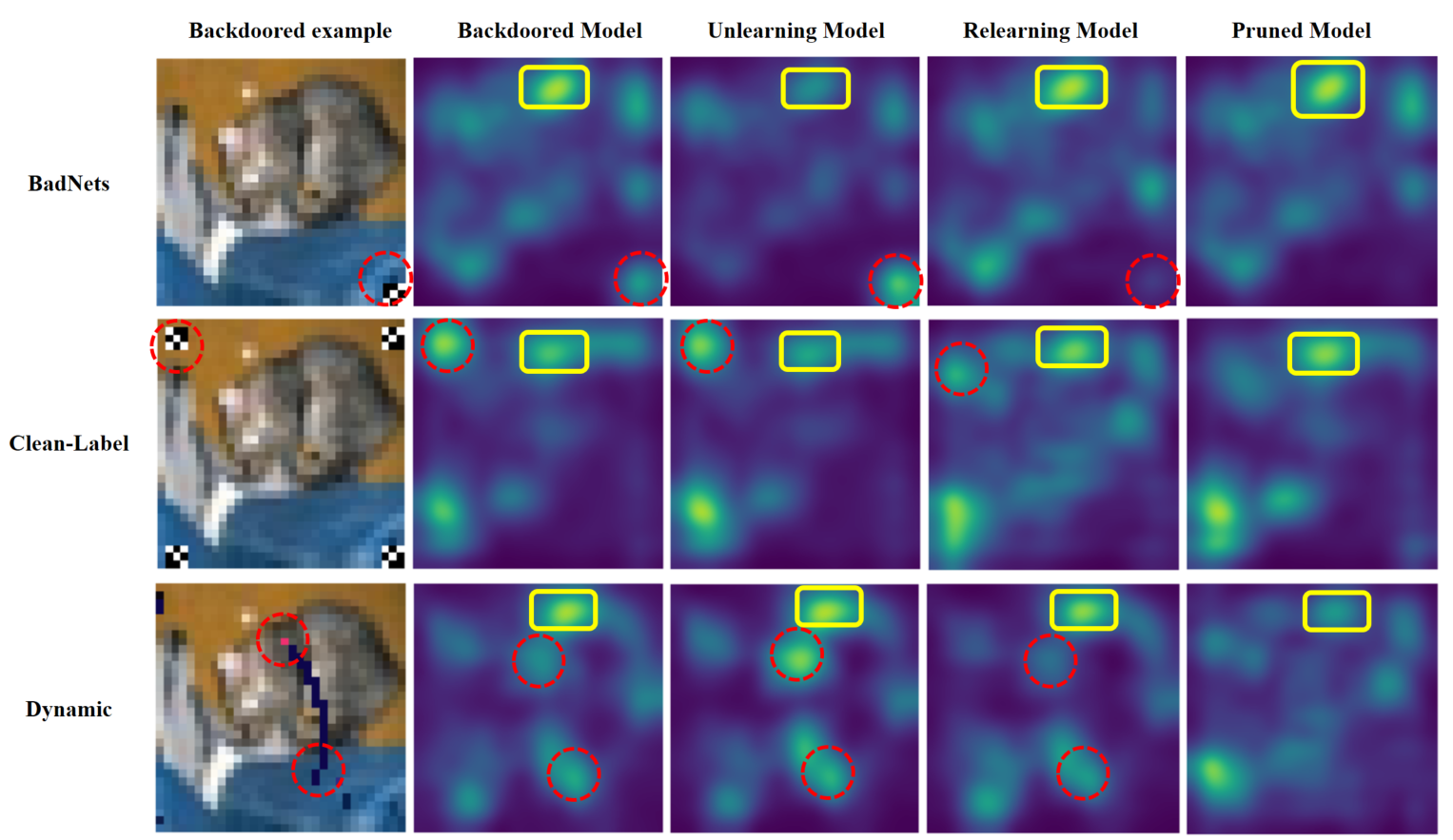}}
\caption{The feature maps (channel-wise averaged) of one backdoored image at the second residual block of a backdoored ResNet-18 model (by BadNets), and the unlearned, the recovered, and the pruned models by our RNP. The rectangles and circles highlight the regions that are mostly activated by the clean and backdoor patterns, respectively.
}
\vskip -0.15in
\label{fig:motivation}
\end{figure}

Figure \ref{fig:motivation} visualizes the feature maps of a backdoored image at the second residual block of a backdoored ResNet-18 model. Specifically, at the unlearning step, the clean neurons (features) are unlearned to have much weaker signals (indicated by the blurry feature map of the unlearned model), especially the most salient features (those in the yellow rectangles). At the recovering step, those weakened clean features will be boosted back to their initial intensity, so their corresponding mask values will increase (rather than decrease). Conversely, the backdoor neurons will be repurposed by the recovering process to compensate (or help boost) the weakened clean neurons, so their mask values will be largely decreased (close to zero), and their activations will almost “disappear”. This can be observed from the dark blue region of the \textit{recovered model} at the bottom right corner (marked by the red dashed circle) of the feature map. Note that the mask $\mathbf{m}^{\kappa}$ is an element-wise scaling matrix applied to each filter, as defined in Eqn. \ref{eq:3}.

The above result indicates that the neuron-level unlearning erases the clean feature (although not completely) while keeping the backdoor feature almost unchanged. The recovered model relocates the neurons associated with the backdoor to compensate for the clean feature, and the backdoor is completely removed from the pruned model. More understanding of why asymmetric unlearning and recovering is the key to exposing backdoor neurons can be found in Section \ref{sec:understanding}.

\begin{table*}[tp!]
\renewcommand{\arraystretch}{1.05} 
\centering
\caption{Performances of 5 backdoor defense methods against 12 backdoor attacks. The experiments were done on CIFAR-10 with ResNet-18 and ImageNet-12 subset with ResNet-34 using only 1\% clean defense data. ASR: attack success rate (\%); CA: clean accuracy (\%). The best results are \textbf{boldfaced}.}
\begin{adjustbox}{width=0.98\linewidth}
\begin{tabular}{c|c|cc|cc|cc|cc|cc|cc}
\toprule
\multirow{2}{*}{Datasets} & \multirow{2}{*}{\begin{tabular}[c]{@{}c@{}}Backdoor \\ Attacks\end{tabular}} & \multicolumn{2}{c|}{No Defense} & \multicolumn{2}{c|}{FP} & \multicolumn{2}{c|}{NAD} & \multicolumn{2}{c|}{I-BAU} & \multicolumn{2}{c|}{ANP} & \multicolumn{2}{c}{RNP (Ours)} \\ \cline{3-14} 
 &  & ASR & CA & ASR & CA & ASR & CA & ASR & CA & ASR & CA & ASR & CA \\ \midrule
\multirow{13}{*}{CIFAR-10} & BadNets & 100.00 & 93.40 & 15.11 & 88.15 & 1.20 & 90.64 & 15.50 & 91.18 & 0.53 & 91.61 & \textbf{0.20} & \textbf{92.22} \\
 & Trojan & 99.90 & 93.15 & 56.51 & 85.43 & 5.68 & 88.72 & 12.78 & 90.46 & \textbf{1.00} & 92.37 & 2.23 & \textbf{92.56} \\
 & Blend & 100.00 & 93.10 & 68.22 & 85.21 & 10.92 & 89.77 & 1.62 & 90.16 & 0.50 & 92.31 & \textbf{0.33} & \textbf{92.62} \\
 & CL & 100.00 & 94.84 & 24.38 & 88.75 & 13.17 & 89.76 & 23.12 & 88.82 & 15.20 & 92.86 & \textbf{8.87} & \textbf{93.12} \\
 & SIG & 90.86 & 94.59 & 19.16 & 87.88 & 0.64 & 89.40 & 29.32 & 89.67 & 1.19 & 92.97 & \textbf{0.43} & \textbf{94.62} \\
 & Dynamic & 99.97 & 91.36 & 41.73 & 83.28 & 13.60 & 88.64 & 18.74 & 86.87 & \textbf{9.20} & 89.66 & 15.24 & \textbf{90.18} \\
 & WaNet & 99.10 & 93.67 & 68.92 & 86.35 & 17.46 & 82.41 & 23.18 & 87.38 & 13.14 & 92.64 & \textbf{10.98} & \textbf{92.83} \\
 & FC & 100.00 & 94.67 & 98.45 & 87.62 & 36.07 & 88.02 & 17.93 & 86.75 & 74.75 & 81.97 & \textbf{1.80} & \textbf{90.93} \\
 & DFST & 100.00 & 94.52 & 88.78 & 85.32 & 12.70 & 88.72 & 25.58 & 87.44 & 10.80 & 90.66 & \textbf{4.61} & \textbf{92.78} \\
 & AWP & 94.39 & 94.30 & 23.17 & 86.04 & 1.71 & 89.13 & 8.71 & 89.62 & \textbf{0.67} & 92.64 & 1.04 & \textbf{94.24} \\
 & LIRA & 100.00 & 92.71 & 87.78 & 83.12 & 32.12 & 86.73 & 51.33 & 82.56 & 20.25 & 87.78 & \textbf{13.51} & \textbf{92.26} \\
 & A-Blend & 71.86 & 92.16 & 85.22 & 81.82 & 18.51 & 85.23 & 33.38 & 85.91 & 23.71 & 90.16 & \textbf{1.09} & \textbf{90.38} \\ \cline{2-14} 
 & Average & 96.34 & 93.54 & 56.45 & 85.75 & 13.65 & 88.10 & 21.77 & 88.07 & 14.25 & 90.64 & \textbf{5.03} & \textbf{92.18} \\ \hline
\multirow{6}{*}{ImageNet-12} & BadNets & 100.00 & 88.53 & 91.70 & 83.23 & 9.12 & 83.26 & 15.38 & 85.15 & 10.25 & 85.21 & \textbf{5.80} & \textbf{85.83} \\
 & Trojan & 100.00 & 89.79 & 93.69 & 81.40 & 12.31 & 82.52 & 19.61 & 84.11 & 7.48 & 87.41 & \textbf{0.59} & \textbf{89.30} \\
 & Blend & 99.90 & 89.44 & 92.14 & 82.13 & 28.76 & 82.93 & 9.34 & 82.27 & 6.21 & 86.40 & \textbf{5.54} & \textbf{86.89} \\
 & SIG & 73.78 & 88.18 & 87.82 & 81.27 & 21.15 & 83.31 & 29.23 & 81.57 & 25.53 & 52.52 & \textbf{15.20} & \textbf{84.15} \\
 & FC & 95.77 & 88.95 & 90.52 & 79.36 & 31.43 & 81.56 & 38.51 & 79.33 & 42.69 & 53.01 & \textbf{17.23} & \textbf{83.36} \\ \cline{2-14} 
 & Average & 93.89 & 88.98 & 91.17 & 81.48 & 20.55 & 82.72 & 22.41 & 82.49 & 18.43 & 72.91 & \textbf{8.87} & \textbf{85.91} \\ \bottomrule
\end{tabular}
\end{adjustbox}
\label{tab:2}

\end{table*}

\section{Experiments}\label{sec:experiments}
\subsection{Experimental Setup}
\textbf{Attack Setup.} We primarily consider 12 state-of-the-art backdoor attacks. These include 7 input-space attacks: BadNets \cite{gu2017badnets}, Trojan \cite{liu2018trojaning}, Blend \cite{chen2017targeted}, Dynamic \cite{nguyen2020input}, WaNet \cite{nguyen2021wanet}, SIG \cite{barni2019new}, and CL \cite{turner2019clean}, as well as 3 feature-space attacks: FC \cite{shafahi2018poison}, DFST \cite{cheng2021deep}, and AWP \cite{garg2020can}. In addition, two recently proposed adaptive attacks termed LIRA \cite{doan2021lira} and Adaptive-Blend (A-Blend) \cite{qi2022circumventing} are also considered. We follow the default settings suggested in their original papers and the open-source codes for most attacks, including the trigger pattern, trigger size, and backdoor label. As in previous works \cite{wu2021adversarial, li2021anti}, we evaluate the defense performance against the 12 attacks on the CIFAR-10 dataset and an ImageNet-12 dataset. The backdoor label of all attacks is set to class 0. The detailed settings of these attacks and datasets are summarized in Table \ref{tab:attacks_overview}. It should be noted that some of the attacks are not considered for ImageNet-12 because 1) they were not proposed for ImageNet and 2) they failed to reproduce on ImageNet-12.

\textbf{Defense Setup.} We consider a total of 8 backdoor defense methods, which include 2 backdoor detection methods: Neural Cleanse (NC) \cite{wang2019neural} and STRIP \cite{gao2019strip}, 5 existing backdoor removal methods: Fine-pruning (FP) \cite{liu2018fine}, Neural Attention Distillation (NAD) \cite{li2021neural}, Adversarial Unlearning of Backdoors via Implicit Hypergradient (I-BAU) \cite{zeng2021adversarial}, Adversarial Neuron Perturbation (ANP) \cite{wu2021adversarial}, Anti-backdoor Learning (ABL) \cite{li2021anti}, and lastly, our RNP. All defenses share limited access to only 500 clean samples as their defense data (for both CIFAR-10 and ImageNet-12). Two typical data augmentation techniques (horizontal flip and random crop) are applied. We follow the open-source codes of FP, NAD, and ANP, and adjust their hyper-parameters to obtain the best performance on different attacks. For our RNP, the maximum unlearning epochs are set to 20 and the actual unlearning epochs range from 5 to 15, depending on the datasets, models, and attacks.

\textbf{Evaluation Metrics.}
We adopt three metrics for defense evaluation: 1) Detection Rate (DR), which is the success rate of the defense in identifying the backdoor label or the backdoored model; 2) Clean Accuracy (CA), which is the model's accuracy on clean test data; and 3) Attack Success Rate (ASR), which is the model's accuracy on backdoored test data.

\begin{table*}[!thp]
\renewcommand{\arraystretch}{1.05}
\renewcommand\tabcolsep{2.5pt}
\small
\centering
\caption{The number of neurons pruned by ANP and our RNP against 12 types of backdoor attacks on the CIFAR-10 dataset. $Neurons \downarrow$ indicates the total number of neurons being pruned, and \textit{on All} means the pruning across all the blocks of ResNet-18. }
\vskip 0.1in
\begin{tabular}{c|c|cccccccccccc}
\toprule
Defense & Metric & BadNets & Trojan & Blend & CL & SIG & Dynamic & WaNet & FC & DFST & AWP & LIRA & A-Blend\\ \midrule
\multirow{3}{*}{ANP} & $Neurons \downarrow$ (on All) & 94 & 96 & 42 & 135 & 88 & 69 & 126 & 199 & 165 & 56 & 158 & 96\\ 
 & ASR (\%) & 0.53 & 1.00 & 0.50 & 15.20 & 1.19 & 9.20 & 13.14 & 74.75 & 10.80 & 0.67 & 20.25 & 23.71\\
 & CA (\%) & 91.61 & 92.37 & 92.31 & 92.86 & 92.97 & 89.66 & 92.64 & 81.97 & 90.66 & 92.64 & 87.78 & 90.16\\ \hline
\multirow{3}{*}{RNP} & $Neurons \downarrow$ (on All) & \textbf{41} & \textbf{48} & \textbf{28} & \textbf{103} & \textbf{73} & \textbf{59} & \textbf{92} & \textbf{155} & \textbf{83} & \textbf{40} & \textbf{112} & \textbf{78} \\  
 & ASR (\%) & 0.20 & 2.23 & 0.33 & 8.87 & 0.43 & 15.24 & 10.98 & 1.80 & 4.61 & 1.04 & 13.51 & 1.09\\
 & CA (\%) & 92.22 & 92.56 & 92.62 & 93.12 & 94.62 & 90.18 & 92.83 & 90.93 & 92.78 & 94.24 & 92.96 & 90.38\\
 \bottomrule
\end{tabular}
\label{tab:pruning_number}
\end{table*}

\subsection{Main Defense Results}
\textbf{Results on CIFAR-10.} Table \ref{tab:2} reports the defense performances of 5 backdoor defense methods against the 12 backdoor attacks on CIFAR-10. It is clear that our RNP achieves the best defense performance by reducing the average ASR from 96.34\% to 5.03\% with the minimum drop of CA (less than 2\% on average). In contrast, FP, NAD, I-BAU, and ANP only reduce the average ASR to 56.45\%, 13.65\%, 21.77\%, and 14.25\%, respectively.

We notice that each defense method has its limitations against some specific attacks. For instance, even though RNP has the best overall performance, it is weaker than ANP in defending against the Dynamic attack by approximately 6.04\% of ASR. On the other hand, while ANP achieves considerable results against most attacks, it has much poorer performance on FC, reducing the ASR only to 74.75\%. We speculate that the adversarial perturbation in ANP cannot effectively locate the backdoor neurons when the backdoor features were optimized to collide with the clean features (and so as the neurons) by the FC attack. NAD also mirrors a similar pattern, which struggles to defend against FC, WaNet, and DFST. Meanwhile, our RNP also outperforms the recent defense I-BAU on all attacks. In terms of ASR, RNP surpasses I-BAU by more than 15\% against BadNets, CL, SIG, and FC attacks, more than 10\% against WaNet, DFST, and more than 1\% against Blend and Dynamic. In terms of CA, RNP outperforms I-BAU by more than 3\% against SIG, Dynamic, WaNet, and FC. Finally, FP has the poorest overall performance with an average ASR higher than 50\% against most attacks, indicating that pruning the most dormant neurons on the clean inputs is not accurate enough against advanced attacks. More results of our RNP with different DNN architectures can be found in Appendix \ref{ap:model-arch}.

\textbf{Results on ImageNet-12.} The results on the ImageNet-12 dataset are also presented in Table \ref{tab:2}. It is evident that our RNP surpasses all 4 baselines in terms of either ASR or CA. Particularly, RNP outperforms FP, NAD, I-BAU, and ANP by 82.3\%, 11.68\%, 13.54\%, and 9.56\% more ASR reduction respectively, with only $\leq 3\%$ decline in CA. Note that the two current SOTA defenses I-BAU and ANP failed to defend against the SIG and FC attacks to a large extent. More specifically, I-BAU/ANP only reduces the ASR to 29.23\%/25.53\% against SIG and 38.51\%/42.69\% against FC attacks. Our RNP is more effective against these two attacks than I-BAU and ANP, decreasing their ASRs to 15.20\%, and 17.23\%, respectively. This is somewhat unsurprising that the two attacks become more effective on high-resolution images, as SIG adds large and repetitive sinusoidal patterns into the background while FC directly attacks the representation space.

In summary, our RNP achieves superior performance against a wide range of attacks compared to the 4 baselines. This is due to RNP's ability to expose and identify backdoor neurons via its asymmetric unlearning-recovering process, leading to a more precise pruning effect. The sensitivity of RNP to different model architectures and poisoning rates are provided in Appendix \ref{ap:model-arch} and Appendix \ref{ap:poisoning-rate}, respectively.

\textbf{The Number of Neurons Pruned by RNP.} 
Here, we show the number of neurons pruned by ANP and our RNP against 12 backdoor attacks and report the results on the CIFAR-10 dataset in Table \ref{tab:pruning_number}. These results show that 1) only a few neurons in a backdoored DNN are responsible for the backdoor functionality; and 2) complex attacks such as CL, WaNet, FC, and LIRA tend to create more backdoor neurons, and in this case, ANP has to prune more neurons than our RNP to reduce the ASR. For instance, ANP needs to prune 94, 96, and 42 neurons to defend against input-space attacks BadNets, Trojan, and Blend, respectively. By contrast, our RNP can achieve comparable or even better defense performance by pruning roughly half the number of neurons. A similar advantage of RNP can also be observed against other advanced attacks. The above findings shed new light on the working mechanism of backdoor attacks and also verify the preciseness of our RNP in locating the backdoor neurons. The results of RNP under different pruning strategies can be found in Appendix \ref{ap:pruning_type}.

\textbf{RNP against Strong Adaptive Attacks.} 
Here, we test the resistance of our RNP defense to strong adaptive attacks. We design two adaptive attacks: 1) Adaptive-distillation, which leverages knowledge distillation to align the neuron activation of a backdoored student network with that of a cleanly trained teacher network; and 2) Adv-training, which adversarially perturbs the backdoor neurons identified by our RNP and then fine-tunes the model on the clean subset of defense samples. Both attacks are designed to make exposing backdoor neurons very difficult. Table \ref{tab:adaptive_attack} reports the defense results of ANP and our RNP against the 2 adaptive attacks. It is clear that ANP has failed to defend against the 2 adaptive attacks to some extent (ASR$=$24.81\% and 54.01\%). Our RNP, on the contrary, is fairly robust to both attacks with the ASR reduced to 13.22\% and 18.09\% respectively while maintaining a high CA.

In the future, it is certainly possible that more advanced attacks could break our defense. However, in the current literature, we believe our RNP, as a simple and general framework, has proven itself to be the most effective defense against the most diverse attacks. We have also tested RNP's effectiveness against all-to-all attacks in Appendix \ref{ap:all2all} and various trigger sizes in Appendix \ref{ap:trigger-size}.

\begin{figure*}[tp!]
    \centering
    \begin{minipage}[t]{0.49\textwidth}
    \centering
    \includegraphics[width = .49\linewidth]{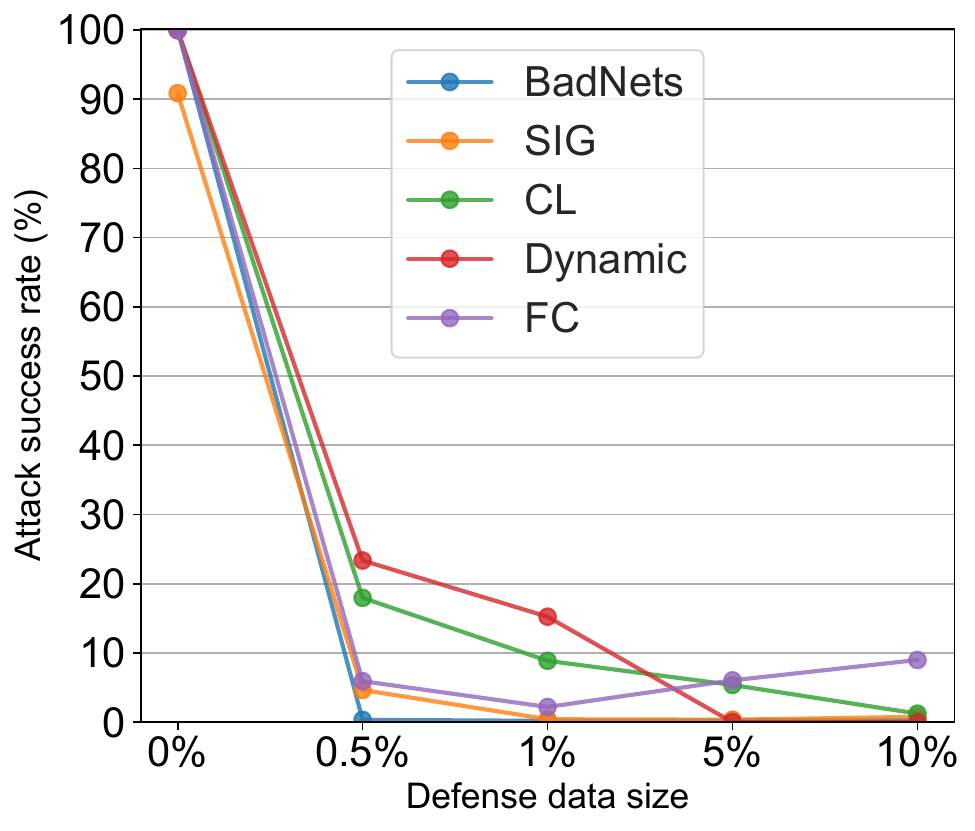}
    \includegraphics[width = .49\linewidth]{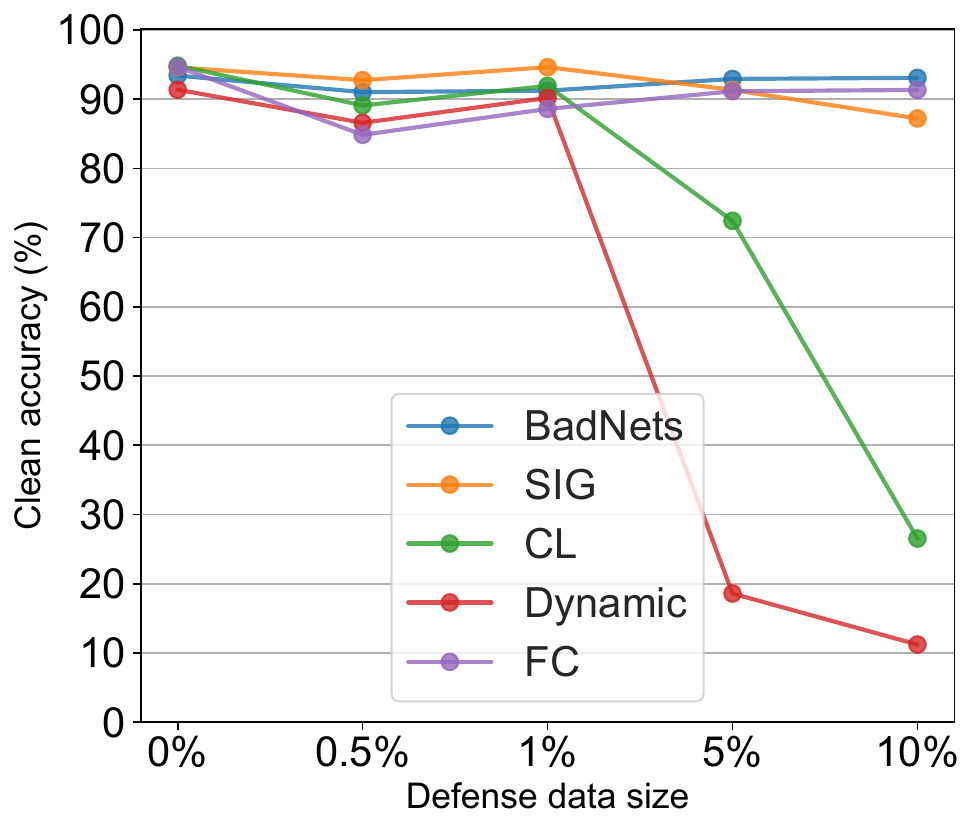}
    \caption{Performance (ASR/CA) of RNP under \emph{different defense data sizes} (\%) against 5 attacks.}
    \label{fig:DF_size}
    \end{minipage}
    \hfill
    \begin{minipage}[t]{0.49\textwidth}
    \centering
    \includegraphics[width = .49\linewidth]{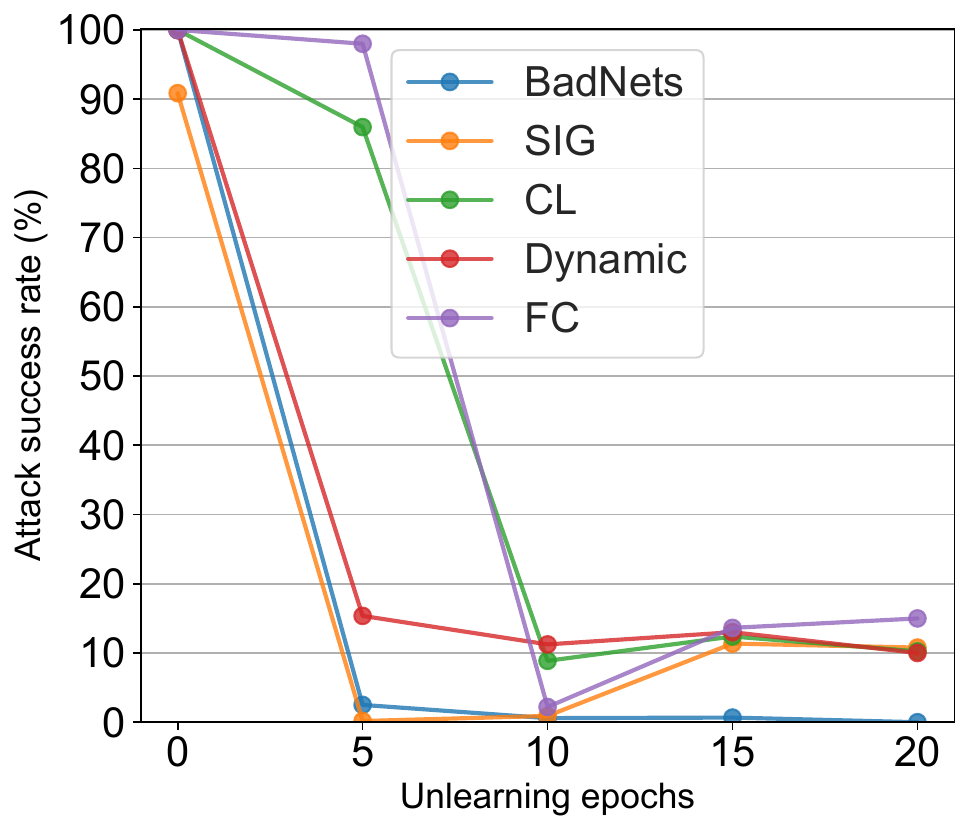}
    \includegraphics[width = .49\linewidth]{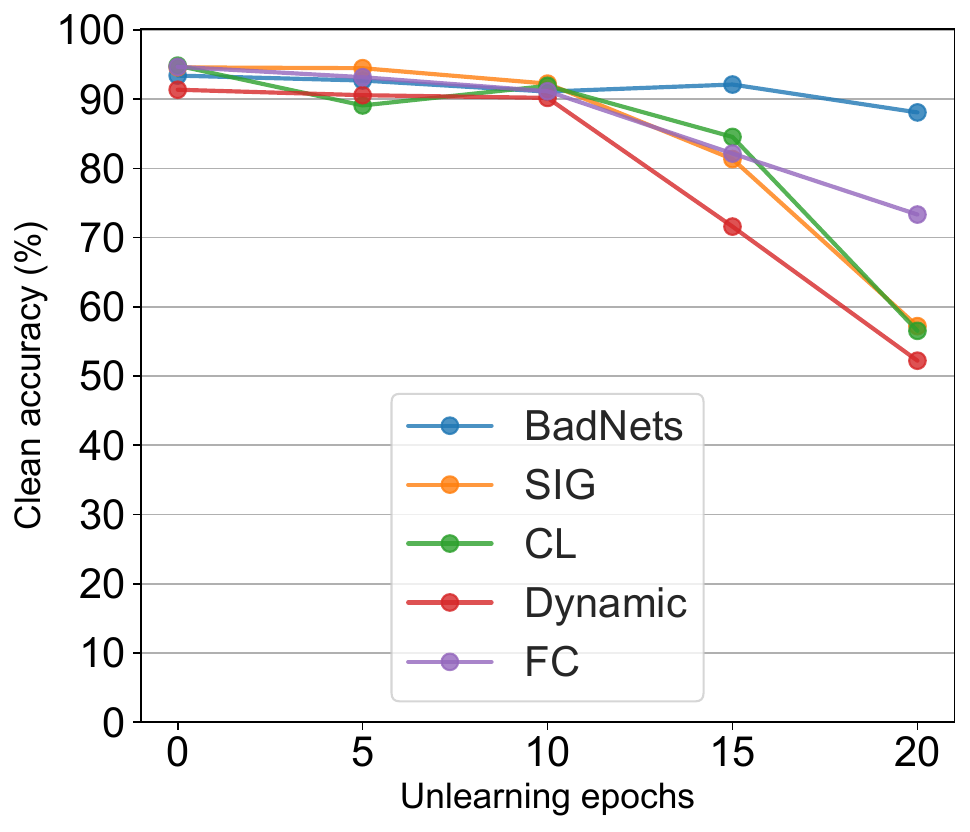}
    \caption{Performance (ASR/CA) of RNP under \emph{different unlearning epochs} against 5 attacks.}
     \label{fig:NU_epoch}
    \end{minipage}
\end{figure*}

\textbf{Impact of the Defense Data Size.}
In this part, we explore the impact of the size of the defense data used to unlearn the backdoored model. We monitor the performance of RNP under varying sizes of the defense (unlearning) set, including 0.5\% (250), 1\% (500), 5\% (2500), and 10\% (5000) of the original clean training set, respectively, and present the results in Figure \ref{fig:DF_size}. We find that the more clean data used in unlearning, the stronger RNP's ability to reduce the ASR for most attacks. However, the results are counter-intuitive for CA, especially for the Dynamic and CL attacks. Surprisingly, the CA drops as the size of the unlearning set grows. More specifically, the CA drops rapidly from nearly 90\% to less than 30\% as we increase the size of the defense set to 10\%. We conjecture that this is because both CL and Dynamic cover the whole image with complex trigger patterns (possibly confusing with the clean feature representation), and thus the more clean data used in unlearning, the higher the chance for RNP to accidentally prune clean neurons. To summarize, our RNP achieves a higher CA with fewer data (i.e., 0.5\% and 1\%) for all these attacks, which is more practical for data-limited situations.

\textbf{Impact of Unlearning Epochs.}
We are also interested in the impact of unlearning epochs on the performance of RNP. Therefore, we investigate the defense performances of RNP against 5 backdoor attacks (i.e., BadNets, SIG, CL, Dynamic, and FC) after the 5th, 10th, 15th, and 20th epochs of unlearning, respectively, and plot them in Figure \ref{fig:NU_epoch}. We find that unlearning the model for an excessive number of epochs hinders defense performance. More specifically, as we increase the unlearning epoch from the 10th to the 15th, the ASR does not drop further (in fact, it even becomes slightly higher), while the CA declines substantially for 4 of the 5 attacks. A possible explanation is that a gradient explosion occurs when maximizing the loss on the clean defense data, causing the unlearned model to collapse in later epochs (e.g., the 15th). This is why we terminate the unlearning immediately when the CA drops to a random guess (i.e., 10\%).

\begin{figure*}[!tp]
\small
\centering
\centerline{\includegraphics[width = .95\linewidth]{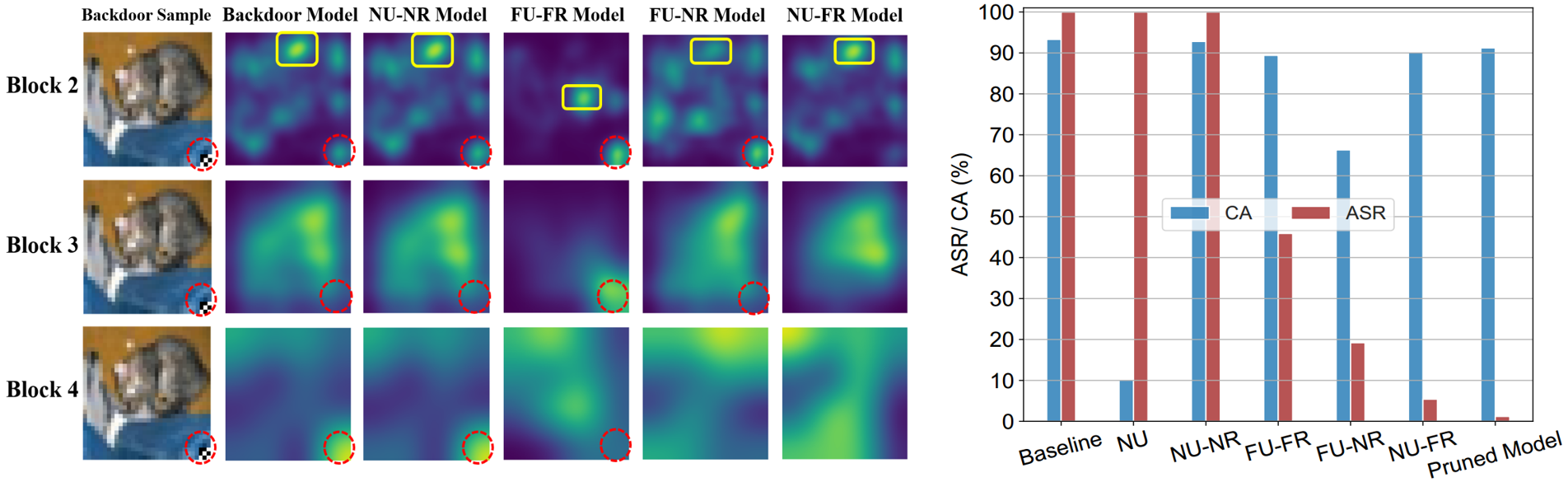}}
\vskip -0.15in
\caption{\textbf{Left:} The feature maps (channel-wise averaged output at the 2nd, 3rd, and 4th residual blocks of ResNet-18) of a backdoored image under 4 different unlearning-recovering strategies: NU-NR, FU-FR, FU-NR, and NU-FR. The squares and circles  highlight the backdoored and clean regions, respectively.  \textbf{Right:} The defense performance (ASR and CA) of the 4 unlearning-recovering strategies. These experiments were conducted with ResNet-18 and BadNets attack on CIFAR-10. \textbf{NU}: neuron unlearning; \textbf{FU}: filter unlearning; \textbf{NR}: neuron recovering; \textbf{FR}: filter recovering.}
\label{fig:attention}
\end{figure*}

\begin{table}[!tp]
\small
\centering
\caption{Defense results of ANP and our RNP on CIFAR-10 dataset against 2 adaptive attacks: 1) Adaptive-distillation (Adapt-D) and 2) Adv-training (Adv-T). }
\vskip 0.1in
\label{tab:adaptive_attack}
\begin{adjustbox}{width=0.95\linewidth}
\begin{tabular}{@{}c|cc|cc|cc@{}}
\toprule
\multirow{2}{*}{\begin{tabular}[c]{@{}c@{}}Adaptive \\ Attacks\end{tabular}} & \multicolumn{2}{c|}{No Defense} & \multicolumn{2}{c|}{ANP} & \multicolumn{2}{c}{RNP (Ours)} \\ \cmidrule(l){2-7} 
        & CA    & ASR   & CA    & ASR   & CA    & ASR   \\ \midrule
Adapt-D & 91.56 & 84.76 & 89.78 & 24.81 & \textbf{90.58} & \textbf{13.22} \\ \midrule
Adv-T & 88.56 & 73.36 & 83.06 & 54.01 & \textbf{86.15} & \textbf{18.09} \\ \bottomrule
\end{tabular}
\end{adjustbox}
\vskip -0.15in
\end{table}

\subsection{Understanding the Mechanism of RNP} \label{sec:understanding}
\textbf{Asymmetric Unlearning-Recovering.} We first show that asymmetric unlearning and recovering are key to successfully exposing backdoor neurons. We demonstrate this by investigating the deep features obtained via different combinations of neuron unlearning (NU), neuron recovering (NR), filter unlearning (FU), and filter recovering (FR).

As can be observed in the left subfigure of Figure \ref{fig:attention}, symmetric unlearning-recovering, i.e., NU-NR and FU-FR, can hardly affect the backdoor features. Neuron recovering in NU-\textbf{NR} and FU-\textbf{NR} can recover more clean features than filter recovering, as it offers more capacity to recover the neurons. Filter recovering (i.e., NU-\textbf{FR} and FU-\textbf{FR}), on the other hand, leads to darker feature maps (fewer activations) at the shallow layers (i.e., Block 2 and Block 3). This implies that filter recovering indeed limits the capacity for clean feature recovery.

Filter unlearning followed by a fine-grained neuron recovering (i.e., FU-NR) can mitigate the backdoor at the deep layers (Block 4) but not at the shallow layers (Block 2). The clean features are largely recovered, yet some backdoor features still exist. Only the neuron unlearning followed by a coarse-grained filter recovering (i.e., NU-FR) can effectively force the model to erase the backdoor features (and the associated backdoor neurons). The defense performance of the 4 strategies shown in the right subfigure of Figure \ref{fig:attention} also confirms its effectiveness in locating and removing the backdoor neurons. In Appendix \ref{ap:filter_location}, we also show that filter recovering can be effectively achieved at a few layers, e.g., layers before each batch normalization. However, it has to be applied across the network at both shallow and deep layers to achieve the best result.

\textbf{Necessity of Unlearning and Recovering.} Apart from the “asymmetric” working mechanism of RNP, we have also run 3 experiments (with CIFAR-10, ResNet-18, and 500 clean images as defense data) to show whether we need both neuron unlearning (maximization) and filter recovering (minimization) in Table \ref{tab:recovering_unlearning} (Appendix \ref{ap:necessity_unlearning}). The findings can be summarized as follows. 1) Unlearning without recovering causes CA to drop significantly, which verifies the usefulness of the recovery step. 2) recovering without unlearning cannot decrease ASR, which proves that unlearning is a MUST. 3) Unlearning by learning incorrectly (i.e., minimizing towards a wrong target rather than maximizing) is notably worse than our RNP. The detailed discussion and analysis of the above findings are provided in Appendix \ref{ap:c}.

\textbf{More Explorations with the Unlearned Model.} Here, we demonstrate one important benefit of our RNP, that is, the unlearned model (NU-model) produced at its unlearning step can be leveraged to improve other backdoor defense tasks. First of all, the unlearned model directly exposes the backdoor label (class) since the functionality of the clean classes has been unlearned. When applying the trigger recovery and backdoor detection method Neural Cleanse (NC) \cite{wang2019neural} on the unlearned model (rather than the original model), one can expose the potential backdoor target more easily and improve the quality of the recovered triggers. When applying the backdoor sample detection method STRIP \cite{gao2019strip} on the unlearned model, one can detect backdoor samples that are notably more complex and stealthy. Moreover, the unlearned model can also help boost the defense performance of existing backdoor removal methods like Fine-Pruning \cite{liu2018fine}. More detailed analyses can be found in Appendix \ref{ap:d}.

\section{Limitation}
While our RNP achieves promising results against existing attacks, it does have several limitations: 1) It is not theoretically guaranteed to defend against all unforeseen attacks, thus having the risk of being compromised by more advanced attacks; 2) RNP faces a noticeable challenge when defending against backdoor attacks with low poisoning rates (e.g., the poisoning rate $\leq$ 1\%), in which backdoor-related neurons/features are much easier to hide within clean neurons/features, leading to inaccurate pruning; 3) RNP fails to erase backdoors trained on lightweight model architectures like EfficientNet, sacrificing too much CA to reduce the ASR without fine-tuning. We will address these limitations of our RNP defense in our future work.

\section{Conclusion}
This paper proposes a novel and effective method called \textit{Reconstructive Neuron Pruning} (RNP) to expose and prune the backdoor neurons from backdoored DNNs. At the core of RNP is an asymmetric unlearning-recovering scheme that first unlearns the neurons on a few clean samples via a \emph{neuron-level unlearning}, and then recovers the neurons on the same clean samples via a \textit{filter-level recovering}. We revealed the phenomenon that asymmetric unlearning-recovering from fine-grained unlearning to coarse-grained recovering can help expose backdoor neurons. We empirically demonstrated the effectiveness of RNP as a backdoor defense method and the benefit of the unlearning technique itself to other backdoor defense tasks, including trigger recovery, backdoor label detection, and backdoor sample detection. We hope our work could provide a new perspective for the community to develop more powerful backdoor defenses in the future.


\section*{Acknowledgements}
This work is supported in part by China National Science Foundation under grant numbers 62072356 and 62276067, and in part by China 111 Project (No. B16037).


\bibliography{example_paper}
\bibliographystyle{icml2023}

\newpage
\appendix
\onecolumn
\section{Implementation Details of RNP} \label{ap:a}
\subsection{Datasets and Classifiers} \label{ap:dataset_classfier}
The datasets and DNN models used in our experiments are summarized in Table \ref{tab:dataset_model}.

\begin{table}[!htbp]
\centering
\caption{Detailed information of the datasets and classifiers used in our experiments.}
\label{tab:dataset_model}
\begin{tabular}{ccccc}
\toprule
Dataset & Labels & Input Size & Training Images & Classifier \\  \midrule
CIFAR-10 & 10 & 32 x 32 x 3 & 50000 & ResNet-18/ VGG-16/ MobileNet-V2  \\
ImageNet subset & 12 & 224 x 224 x 3 & 12406 & ResNet-34 \\
\bottomrule
\end{tabular}
\end{table}

\subsection{Attack Details} \label{ap:attack_details}
We mainly considered 12 state-of-the-art backdoor attacks, including 7 input-space attacks: BadNets \cite{gu2017badnets}, Trojan \cite{liu2018trojaning}, Blend \cite{chen2017targeted}, Dynamic \cite{nguyen2020input}, WaNet \cite{nguyen2021wanet}, SIG \cite{barni2019new}, and CL \cite{turner2019clean}, and 3 feature-space attacks: FC \cite{shafahi2018poison}, DFST \cite{cheng2021deep}, and AWP \cite{garg2020can}. In addition, we evaluated two recently proposed adaptive attacks termed LIRA \cite{doan2021lira} and Adaptive-Blend (A-Blend) \cite{qi2022circumventing}.

\begin{figure}[!tp] 
\vskip 0.25in
\small
\centering
\centerline{\includegraphics[width = .80\linewidth]{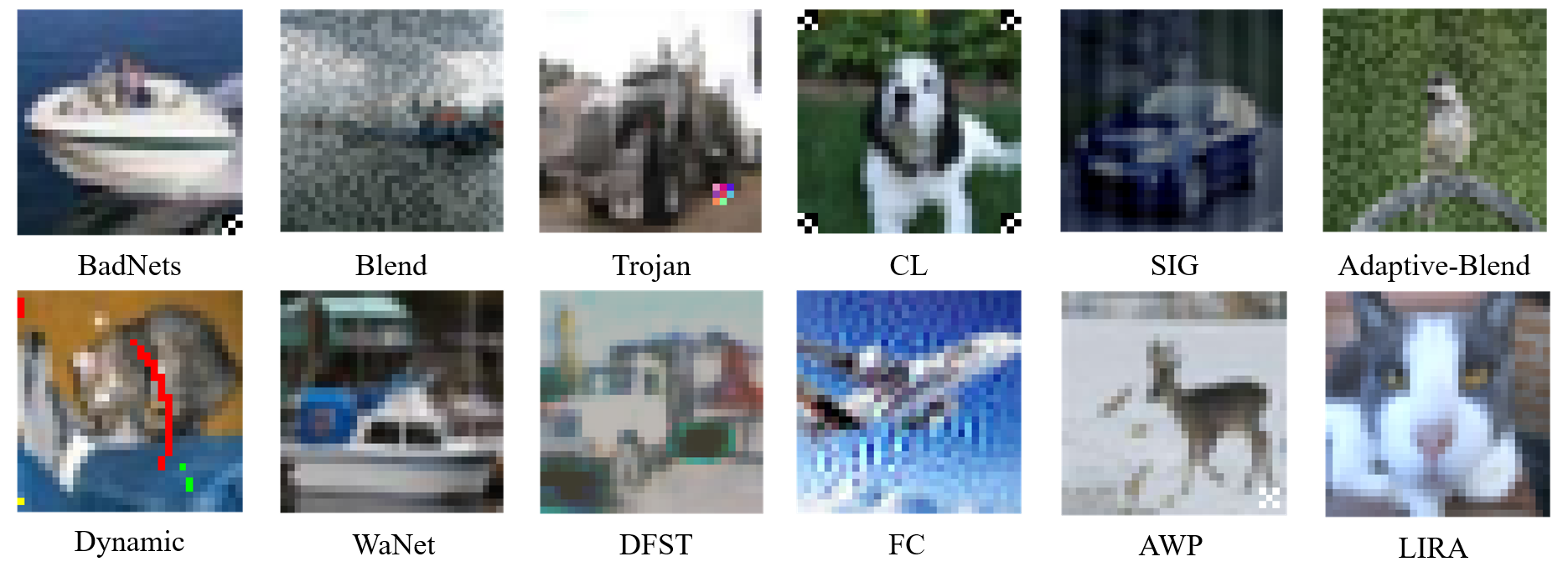}}
\caption{Examples of 12 backdoor trigger patterns on CIFAR-10.}
\label{fig:backdoor_example}
\end{figure}

Figure \ref{fig:backdoor_example} shows a few examples of the backdoor triggers used in our experiments. All attacks were mainly trained on CIFAR-10 \cite{krizhevsky2009learning} to attack the ResNet-18 model \cite{he2016deep} or an ImageNet-12 \cite{deng2009imagenet} subset to attack ResNet-34. We trained all models for 200 epochs using Stochastic Gradient Descent (SGD) with an initial learning rate of 0.1, a batch size of 128, and a weight decay of 5e-4 to obtain the backdoored models. The learning rate was divided by 10 at the 60th and 120th epochs. We used two standard data augmentation techniques (horizontal flip and random crop with padding $4 \times 4$) during model training. We followed the default settings suggested in the original papers and the open-source codes for most attacks; this included the trigger pattern, trigger size, and backdoor label. We also tuned the hyperparameters of several attacks that were negatively affected by the two data augmentations to obtain the best attack performance. We carefully altered the hyperparameter configurations for several feature-space attacks to ensure that they achieved the best attack performances. The backdoor label of all attacks was set to class 0. We also evaluated the defense performance of our RNP on an ImageNet-12 subset. Following previous work \cite{li2021anti}, we reproduced 5 attacks on ImageNet-12: BadNets, Blend, Trojan, SIG, and FC. We omitted the other attacks due to failed reproductions. Table \ref{tab:attacks_overview} summarizes the detailed settings of these attacks.

\begin{table}[!htbp]
\renewcommand{\arraystretch}{1.05} 
\centering
\caption{Attack settings of 12 backdoor attacks. ASR (\%): attack success rate; CA (\%): clean accuracy.}
\label{tab:attacks_overview}
\begin{tabular}{ccccc}
\toprule
Attacks & Trigger Type & Trigger Pattern & Target Label & Poisoning Rate \\ 
\midrule
BadNets & Fixed & Grid & 0 & 0.1 \\
Trojan & Fixed & Reversed Watermark & 0 & 0.1  \\
Blend & Fixed & Random Pixel & 0 & 0.1  \\
Dynamic & Varied & Mask Generator & 0 & 0.1 \\
SIG & Fixed & Sinusoidal Signal & 0 & 0.08 \\
CL & Fixed & Grid and PGD Noise & 0 & 0.08 \\
FC & Varied & Optimization-based & source 1, target 0 & 0.08 \\
DFST & Varied & Style Generator & 0 & 0.1 \\
WaNet & Varied & Optimization-based & 0 & 0.1 \\
AWP & Fixed & Weight Perturbation & 0 & 0.1 \\
LIRA & Varied & Optimization-based & 0 & 0.1 \\
A-Blend & Fixed & Mixer Construction & 0 & 0.1 \\
\bottomrule
\end{tabular}
\end{table}

\subsection{Defense Details} \label{ap:defense_details}
We experimented with 8 backdoor defenses in total, including 2 backdoor detection methods: Neural Cleanse (NC) \cite{wang2019neural} and STRIP \cite{gao2019strip}, and 5 backdoor removal methods: Fine-pruning (FP) \cite{liu2018fine}, Neural Attention Distillation (NAD) \cite{li2021neural}, Adversarial Unlearning of Backdoors via Implicit Hypergradient (I-BAU) \cite{zeng2021adversarial}, Adversarial Neuron Perturbation (ANP) \cite{wu2021adversarial}, Anti-backdoor Learning (ABL) \cite{li2021anti}, and our RNP. All defenses had limited access to only 500 defense data from the CIFAR-10 training set (or ImageNet-12).

We used the open-source PyTorch code for NC \footnote{https://github.com/VinAIResearch/input-aware-backdoor-attack-release/tree/master/defenses/neural\_cleanse} to reproduce the results of backdoor detection and trigger recovery. To combine our NU with the existing method NC (i.e., NU+NC), we replaced only the original model $f$ used by NC with our unlearned model $f_{\hat{\theta}}$ and kept other settings unchanged. For STRIP, we calculated the relative entropy between the backdoored model's output distributions on clean vs. backdoored examples. We then compared the difference in relative entropies between the original backdoored model and the unlearned backdoored model $\hat{\theta}_{NU}$.

We reimplemented FP using PyTorch and pruned the last convolutional layer (i.e., Layer4.conv2) of the model until the CA of the network became lower than 80\%. For NAD, we adopted the same settings used in the open-sourced code \footnote{https://github.com/bboylyg/NAD} and cautiously selected the best hyper-parameter $\beta$ from $[0, 5000]$ with an interval of 500. For I-BAU, we followed the settings used in the open-sourced code \footnote{https://github.com/YiZeng623/I-BAU} to present the best defense results. We used the open-source code for ANP \footnote{https://github.com/csdongxian/ANP\_backdoor}, and followed the suggested settings with the perturbation budget $\epsilon = 0.4$ and the trade-off coefficient $\alpha = 0.2$ to optimize the mask. We also combined our NU with NC to recover the trigger patterns and then erased the triggers from the backdoored model using the unlearning technique used in ABL.

For our RNP defense, we maximized the unlearned model $f_{\hat{\theta}}$ for 20 epochs until its clean accuracy dropped to 10\% (random guess) with a learning rate of 0.01, a batch size of 128, and a weight decay of 5e-2. For the recovering step, we optimized the filter mask for 20 epochs with a learning rate of 0.2. In comparison to pruning by fraction, we found that pruning the neurons by a dynamic threshold often yielded better performance, and adopting a threshold within $[0.4, 0.7]$ consistently produced the best results of RNP (low ASR and high CA) against all backdoor attacks. Note that ANP also suggested the dynamic threshold strategy. The neurons were pruned based on the learned mask using the dynamic thresholding strategy for an accuracy drop $\sim2\%$. See Table \ref{tab:pruning_type} and Table \ref{tab:pruning_number} for more detailed analyses of different pruning strategies and pruning rates.

All defense methods were trained using the same data augmentation techniques, i.e., random crop ($padding = 4$) and horizontal flipping as mentioned in the attack settings.

\begin{table}[!tp]
\small
\centering
\caption{Comparison between ANP and our RNP against 6 backdoor attacks on GTSRB dataset. Note that only 500 clean samples are used as defense data.}
\label{tab:gtsrb_RNP}
\begin{tabular}{cccccccc}
\toprule
\multicolumn{2}{c}{\multirow{2}{*}{Backdoor Attacks}} & \multicolumn{2}{c}{No Defense} & \multicolumn{2}{c}{ANP} & \multicolumn{2}{c}{RNP (Ours)} \\ \cline{3-8} 
\multicolumn{2}{c}{}        & ASR   & CA    & ASR  & CA    & ASR  & CA    \\ \hline
\multicolumn{2}{c}{BadNets} & 100   & 96.83 & 3.59 & 95.37 & 0.16 & 95.49 \\
\multicolumn{2}{c}{Blend}   & 100   & 96.71 & 2.59 & 94.57 & 1.64 & 95.54 \\
\multicolumn{2}{c}{Trojan}  & 99.97 & 96.14 & 5.73 & 94.27 & 6.96 & 95.35 \\
\multicolumn{2}{c}{CL}      & 68.65 & 95.53 & 5.8  & 91.32 & 0    & 94.24 \\
\multicolumn{2}{c}{SIG}     & 85.13 & 96.33 & 0    & 22.89 & 5.74 & 94.91 \\
\multicolumn{2}{c}{Dynamic} & 100   & 97.11 & 7.27 & 96.58 & 6.11 & 96.65 \\ 
\bottomrule
\end{tabular}
\end{table}

\section{Additional Experimental Results of RNP} \label{ap:b}
\subsection{Comparison between RNP and ANP on the GTSRB Dataset} \label{ap:GTSRB_RNP}

We conducted a set of new experiments on GTSRB dataset. Table \ref{tab:gtsrb_RNP} reports the performance of ANP and our RNP against 6 typical attacks (BadNets, Trojan, Blend, CL, SIG, Dynamic). The results indicate that our RNP method generalizes well and consistently outperforms ANP on this new dataset against the 6 attacks.

\subsection{RNP with Different Pruning Strategies} \label{ap:pruning_type}
Existing work \cite{wu2021adversarial} implemented neuron pruning with two different strategies: threshold or fraction. The threshold-based strategy prunes the neurons with a mask value smaller than the pre-specified threshold (by setting the neuron weight to 0), while the fraction-based strategy prunes a certain proportion of the neurons via the mask. To verify their effectiveness, we trained BadNets models using ResNet-18 with similar settings as described in Section \ref{sec:experiments}. For a fair comparison, we reported the performance metrics such as ASR, CA, and the number of pruned neurons in Table \ref{tab:pruning_type}. We found that the threshold-based strategy achieves better results on both ASR and CA than fraction-based pruning (ASR: 92.22 VS. 91.82, CA: 0.20 VS. 0.47). Interestingly, the superiority of the threshold-based strategy is also reflected by the number of pruned neurons (Neurons: 41 vs. Neurons: 48); fewer pruned neurons result in less effect on the CA. This finding provides another piece of evidence of how effective and precise our RNP can expose the backdoor neurons. As a result, we set the threshold-based strategy as the default setting throughout all experiments unless otherwise specified.

\begin{table}[!htbp] 
\small
\centering
\caption{Comparison to the pruning strategy by \textit{dynamic threshold} and \textit{by fraction} against BadNets on ResNet-18, respectively. $Neurons \downarrow$ indicates the number of neurons pruned.}
\vskip 0.25in
\begin{tabular}{ccccccccc}
\toprule
By Threshold & 0 & 0.2 & 0.5 & 0.6 & 0.65 & 0.7 & 0.75 & 0.8 \\ \hline
$Neurons \downarrow$ & 0 & 0 & 5 & 11 & 23 & \textbf{41} & 75 & 178 \\
ASR (\%) & 100 & 100 & 100 & 95.50 & 81 & \textbf{0.2} & 0 & 0 \\
CA (\%) & 93.4 & 93.4 & 93.39 & 92.84 & 92.8 & \textbf{92.22} & 91.17 & 86.92 \\ \midrule
By Fraction (\%) & 0 & 1 & 2 & 3 & 4 & 5 & 8 & 10 \\ \hline
$Neurons \downarrow$ & 0 & \textbf{48} & 93 & 144 & 192 & 240 & 384 & 432 \\
ASR (\%) & 100 & \textbf{0.47} & 0.16 & 0 & 0 & 0 & 0 & 0 \\
CA (\%) & 93.4 & \textbf{91.82} & 89.81 & 89.02 & 85.71 & 81.68 & 72.20 & 64.93 \\ 
\bottomrule
\label{tab:pruning_type}
\end{tabular}
\end{table}

\subsection{RNP with Different Model Architectures} \label{ap:model-arch}
We also evaluated our RNP with additional model architectures, including ResNet-18, VGG-19, MobileNet-V2, and EfficientNet-B0. Table \ref{tab:model-arch} reports the performance of ANP and our RNP against 3 typical attacks (BadNets, Trojan, Blend) on the CIFAR-10 dataset. The results indicate that our RNP generalizes well across most different model architectures (i.e., ResNet-18, VGG-19, and MobileNet-V2) but faces challenges with the EfficientNet architecture.

Specifically, we found that RNP achieves better defense performance than ANP on ResNet-18, VGG-19, and MobileNet-V2 while facing the same challenge as ANP on EfficientNet-B0, where the accuracy of the pruned model dropped significantly. We speculate that this result may be related to the design of lightweight model structures. The sparse neurons in lightweight model structures limit the asymmetric unlearning-recovering and make it difficult to locate the backdoor neurons accurately, thereby leading to decreased performance of RNP. We will investigate and address the performance degradation of RNP on such lightweight model structures in our future research.

\vspace{-0.1in}

\begin{table}[!htbp]
\vskip 0.15in
\centering
\small
\caption{Defense performance (ASR and CA) of ANP and our RNP with different model architectures on the CIFAR-10 dataset. Note that only 500 images are used for defense.}
\label{tab:model-arch}
\begin{adjustbox}{width=0.9\linewidth}
    \begin{tabular}{c|c|ccc|ccc|ccc|ccc}
    \toprule
    \multirow{2}{*}{Model Architectures} & \multirow{2}{*}{Metric} & \multicolumn{3}{c|}{ResNet-18} & \multicolumn{3}{c|}{Vgg-19} & \multicolumn{3}{c}{MobileNet-V2} & \multicolumn{3}{c}{EfficienNet-B0}\\ \cline{3-14} 
     &  & \begin{tabular}[c]{@{}c@{}}No\\ Defense\end{tabular} & ANP & \begin{tabular}[c]{@{}c@{}}RNP\\ (Ours)\end{tabular} & \begin{tabular}[c]{@{}c@{}}No\\ Defense\end{tabular} & ANP & \begin{tabular}[c]{@{}c@{}}RNP\\ (Ours)\end{tabular} & \begin{tabular}[c]{@{}c@{}}No\\ Defense\end{tabular} & ANP & \begin{tabular}[c]{@{}c@{}}RNP \\ (Ours)\end{tabular} & \begin{tabular}[c]{@{}c@{}}No\\ Defense\end{tabular} & ANP & \begin{tabular}[c]{@{}c@{}}RNP\\ (Ours)\end{tabular} \\ \midrule
    \multirow{2}{*}{BadNets} & ASR & 100 & 0.53 & 0.20 & 100 & 3.31 & 0.38 & 100 & 5.92 & 3.17 & 100 & 3.12 & 2.79 \\ \cline{2-14} 
     & CA & 93.40 & 91.61 & 92.22 & 92.63 & 92.37 & 92.57 & 92.89 & 90.48 & 91.67 & 76.74 & 34.31 & 65.62\\ \hline
    \multirow{2}{*}{Trojan} & ASR & 99.90 & 1.00 & 2.23 & 99.98 & 2.60 & 2.49 & 100 & 2.03 & 2.96 & 100 & 3.63 & 1.18\\ \cline{2-14} 
     & CA & 93.15 & 92.37 & 92.56 & 92.67 & 92.44 & 92.31 & 93.01 & 90.91 & 90.21 & 76.48 & 28.83 & 68.17\\ \hline
    \multirow{2}{*}{Blend} & ASR & 100 & 0.50 & 0.33 & 100 & 2.93 & 2.72 & 100 & 1.79 & 1.21 & 100 & 6.32 & 4.38 \\ \cline{2-14} 
     & CA & 93.10 & 92.31 & 92.62 & 92.60 & 92.17 & 92.06 & 92.80 & 86.71 & 90.26 & 75.52 & 21.43 & 63.77\\ \bottomrule
    \end{tabular}
\end{adjustbox}
\end{table}

\begin{table}[!htbp]
\centering
\caption{Comparison between ANP and our RNP against 3 all-to-all backdoor attacks on CIFAR-10. Note that only 500 clean samples are used as defense data.}
\label{tab:all2all}

\begin{tabular}{c|cc|cc|cc}
\toprule
\multirow{2}{*}{Backdoor Attacks} & \multicolumn{2}{c|}{No Defense} & \multicolumn{2}{c|}{ANP} & \multicolumn{2}{c}{RNP (Ours)} \\ \cline{2-7} 
 & ASR & CA & ASR & CA & ASR & CA \\ \midrule
BadNets (all-to-all) & 92.18 & 92.65 & 6.16 & 88.58 & \textbf{1.24} & \textbf{92.56} \\
Trojan (all-to-all) & 91.98 & 92.57 & 18.84 & 88.14 & \textbf{8.78} & \textbf{90.67} \\
Blend (all-to-all) & 84.23 & 92.38 & 13.81 & 86.26 & \textbf{11.56} & \textbf{90.02} \\\bottomrule
\end{tabular}
\end{table}

\begin{table}[!htbp]
\centering
\caption{Defense performance of our RNP against BadNets under various trigger sizes on CIFAR-10. Note that only 500 clean samples are used as the defense data.}
\label{tab:trigger-size}
\begin{tabular}{c|cc|cc|cc}
\toprule
\multirow{2}{*}{Backdoor Attacks} & \multicolumn{2}{c|}{No Defense} & \multicolumn{2}{c|}{ANP} & \multicolumn{2}{c}{RNP (Ours)} \\ \cline{2-7} 
 & ASR & CA & ASR & CA & ASR & CA \\ \midrule
$3\times3$  & 100 & 93.40 & 0.53 & 91.61 & \textbf{0.20} & \textbf{92.22} \\
$5\times5$  & 100 & 92.94 & \textbf{0.74} & 89.72 & 1.44 & \textbf{91.38} \\
$10\times10$  & 100 & 91.78 & 9.08 & 88.53 & \textbf{1.76} & \textbf{91.66} \\\bottomrule
\end{tabular}
\end{table}

\subsection{RNP against All-to-All Attacks} \label{ap:all2all}
We reproduced 3 typical all-to-all (i.e., target\_label $=$ original\_label $+ 1$) attacks with ResNet-18 on the CIFAR-10 dataset.

Here, we report the defense results for both our RNP and ANP. For ANP, we directly used its open-sourced code and selected its best defense results for comparison. The results, presented in Table \ref{tab:all2all}, show that ANP achieves good defense performance in terms of both ASR and CA. However, our RNP method still outperformed ANP with noticeable margins. For example, the CA of our RNP is all above 90\% while ANP is around 86\% to 88\%. The ASR reduction of our method is $\sim5\%$, $\sim10\%$, and $\sim2\%$ better than ANP against BadNets, Trojan, and Blend attacks, respectively.

\subsection{RNP against Different Trigger Sizes} \label{ap:trigger-size}
The performance of our RNP against BadNets under different trigger sizes is reported in Table \ref{tab:trigger-size}. It shows that our RNP can defend against the attack with different trigger sizes from $3 \times 3$ to $10 \times 10$. Note that a $10 \times 10$ trigger appears rather large and obvious on $32 \times 32$ CIFAR-10 images. Increasing the trigger size does bring slightly more resistance to our defense but is very limited. Please note that, except for BadNets, Trojan, and AWP attacks, the triggers of other attacks span the entire image, that is, they have a fixed trigger size – the image/input size.

\begin{table}[!htbp]
\small
\centering
\caption{Defense performance of ANP and our RNP against BadNets attack on CIFAR-10 with diverse poisoning rates. Note that only 500 images are used for defense.}
\label{tab:poisoning-rate}
\begin{tabular}{cccccccccc}
\toprule
\multicolumn{2}{c}{\multirow{2}{*}{CIFAR-10: BadNets}} &
  \multicolumn{2}{c}{No Defense} &
  \multicolumn{2}{c}{ANP} &
  \multicolumn{2}{c}{RNP} &
  \multicolumn{2}{c}{RNP + Finetuning} \\ \cline{3-10} 
\multicolumn{2}{c}{}              & ASR   & CA    & ASR  & CA    & ASR  & CA    & ASR  & CA    \\ \hline
\multicolumn{2}{c}{10\% (5000)}  & 100   & 93.4  & 0.53 & 91.61 & 0.2  & 92.22 & 3.78 & 92.64 \\
\multicolumn{2}{c}{5\% (2500)}   & 100   & 92.59 & 1.63 & 78.83 & 0.32 & 90.26 & 2.51 & 90.18 \\
\multicolumn{2}{c}{1\% (500)}    & 99.96 & 92.67 & 75.8 & 59.42 & 5.29 & 84.8  & 4.45 & 86.59 \\
\multicolumn{2}{c}{0.05\% (250)} & 99.18 & 92.36 & 0    & 14.42 & 0    & 56.32 & 0.46 & 77.29 \\ \bottomrule
\end{tabular}
\end{table}

\subsection{RNP against Different Poisoning Rate} \label{ap:poisoning-rate}

We evaluated RNP's performance against the BadNets attack on CIFAR-10 with diverse poisoning rates. From Table \ref{tab:poisoning-rate}, we observed that both ANP and RNP inevitably encounter clean accuracy degradation as the poisoning rate decreases. When the poisoning rate is 0.05\%, the clean accuracy of ANP drops to 14.42\%, while our RNP drops to 56.32\% but improves to 77.29\% when fine-tuning is applied. We conjecture that this is because the low poisoning rate hides the backdoor neurons more stealthily among the benign neurons, making pruning more challenging. However, low poisoning rates pose a challenge for most current state-of-the-art backdoor defenses, including backdoor sample detection (e.g., STRIP \cite{gao2019strip}), backdoor model detection (e.g., Neural Cleanse \cite{wang2019neural}), backdoor model/neuron removal (e.g., ANP \cite{wu2021adversarial}), and robust backdoor learning (e.g., ABL \cite{li2021anti}). Effective solutions to such a challenge deserve further investigation.

\begin{table*}[!thp]
\centering
\small
\caption{Defense results of 1) pruning without filter recovering; 2) filter recovering with/without neuron unlearning; and 3) unlearning by learning incorrectly. The experiments are implemented on CIFAR-10 with ResNet-18, and 500 clean images as the defense data.
}
\label{tab:recovering_unlearning}
\begin{tabular}{@{}c|cc|cc|cc|cc|cc@{}}
\toprule
\multirow{2}{*}{\begin{tabular}[c]{@{}c@{}}Pruning\\ Results\end{tabular}} &
  \multicolumn{2}{c|}{No Defense} &
  \multicolumn{2}{c|}{\begin{tabular}[c]{@{}c@{}}Pruning w/o \\ Recovering\end{tabular}} &
  \multicolumn{2}{c|}{\begin{tabular}[c]{@{}c@{}}Filter Recovering \\ w/o Unlearning\end{tabular}} &
  \multicolumn{2}{c|}{\begin{tabular}[c]{@{}c@{}}Filter Recovering \\ w/ Unlearning (RNP)\end{tabular}} &
  \multicolumn{2}{c}{\begin{tabular}[c]{@{}c@{}} Learning Incorrectly\end{tabular}} \\ \cmidrule(l){2-11} 
        & CA    & ASR   & CA    & ASR  & CA    & ASR   & CA    & ASR   & CA    & ASR   \\ \midrule
BadNets & 93.40 & 100   & 40.5  & 0    & 93.39 & 100   & 92.22 & 0.20  & 80.58 & 3.64  \\ \midrule
SIG     & 94.59 & 90.86 & 29.49 & 0.38 & 94.57 & 90.61 & 94.62 & 0.43  & 93.78 & 0.73  \\ \midrule
CL      & 94.84 & 100   & 11.82 & 0.44 & 93.84 & 100   & 91.92 & 8.87  & 71.12 & 4.00  \\ \midrule
Dynamic & 91.36 & 99.97 & 31.2  & 0    & 91.26 & 99.85 & 90.18 & 15.24 & 83.33 & 76.06 \\ \midrule
FC      & 94.67 & 100   & 18.23 & 7.52 & 94.41 & 100   & 90.93 & 1.80  & 80.23 & 17.33 \\ \bottomrule
\end{tabular}
\end{table*}

\section{More Understandings of RNP} \label{ap:c}
\subsection{Necessity of Neural Unlearning and Filter Recovering} \label{ap:necessity_unlearning}
In addition to "asymmetric unlearning-recovering," we have conducted 3 new experiments (with CIFAR-10, ResNet-18, and 500 clean images as defense data) to demonstrate the key element of \textit{neural unlearning and filter recovering}. The conclusions have also been carefully checked against other attacks. The settings and findings of these experiments are summarized below, with the results reported in Table \ref{tab:recovering_unlearning}.

\textbf{\textit{a) Pruning without filter recovering, where the backdoored neurons are directly pruned from the unlearned model. This is to validate the usefulness of the recovering step.}}

\textbf{\textit{Finding:}} The backdoor can be reliably erased from the model (most of the ASR is close to 0), but the clean accuracy (CA) is below 41\%. The low ASR indicates that the backdoor neurons can also be removed even without recovering, but this could also remove a certain amount of clean neurons (CA drops significantly). This can be explained by the visualizations (Unlearned model vs. Recovered model) in Figure \ref{fig:motivation} in our main paper. Without the recovering, the key clean neurons (in the yellow rectangles) will be largely damaged, and some clean neurons outside the rectangles will be changed too slightly. This will cause two negative effects: 1) the clean accuracy is hard to recover (caused by damaged clean neurons), and 2) some clean neurons may accidentally get removed. This also verifies the importance of the recovering step (a comparison of neuron recovering vs. filter recovering has already been analyzed in Figure \ref{fig:attention}). Please note that, without the recovering, the neurons that DO NOT change much during the unlearning process should be removed (as only the clean neurons will be unlearned). However, with recovering, the neurons that change the MOST should be removed (as backdoor neurons are repurposed).

\textbf{\textit{b) Filter recovering with/without neuron unlearning, which is to show that neuron unlearning is a MUST.}}

\textbf{\textit{Finding:}} The result shows that filter recovering alone cannot decrease the ASR (over 90\%) although the clean accuracy is as good as the original model. This is somewhat unsurprising, as filter recovering without unlearning is the same as directly fine-tuning the model on the defense data.

\textbf{\textit{c) Unlearning by learning incorrectly, where the backdoored model is fine-tuned on incorrectly labeled clean samples (the labels are set to be their predicted labels of the RNP-unlearned model). That is, this model optimizes the same objective as our RNP but via a process of minimization (learning incorrectly) rather than maximization (unlearning).}}

\textbf{\textit{Finding:}} This defense is notably worse than our RNP, which indicates that minimization and maximization are quite different in terms of their impact on the neurons. Maximization (unlearning) can be understood from the perspective of adversarial perturbation applied to the neurons, i.e., it updates the most influential neurons to maximize the model's error (as it is the quickest way to reduce the loss). Minimization, on the other hand, tends to update all neurons (even repurpose some of the neurons, as we have shown in Figure \ref{fig:attention}) to reduce the classification loss. We believe this finding itself is interesting even out of the scope of backdoor defense.

\subsection{How Does Filter Location Impact RNP?} \label{ap:filter_location}
Table \ref{tab:filter_location} shows how the location of filter recovering affects the defense performance of our RNP.
We restrict the filter mask to be applied before the last batch normalization layer at different residual blocks of the network and report the defense performance after pruning. It shows that the defense is more effective if the mask is applied at deeper layers (Block 4 vs. Block 1), in terms of both ASR and CA.
This indicates that the attack strength accumulates from shallow to deep layers and becomes more malicious in the deep layers (pointing to the backdoor class). It is worth mentioning that, when only recovering and pruning Block 4, the clean accuracy is largely preserved, or even slightly improved: 93.61\% vs. 93.40\% (in Table \ref{tab:filter_location}). This implies that sweeping the deep layers may be a good option if clean accuracy is the primary concern in real-world applications. RNP becomes most effective when filter recovering is applied across the entire network (All Block), which is also the default setting of our experiments. It achieves the best ASR performance, but the CA is reduced by approximately 2\%. Therefore, we recommend this strategy if security is the primary concern in real-world applications.

\begin{table}[!tp]
\small
\centering
\caption{Filter recovering at different residual blocks of ResNet-18. $Neurons \downarrow$ indicates the number of neurons pruned at the corresponding block of ResNet-18 against BadNets attack.}
\begin{tabular}{cccccc}
\toprule
\multirow{2}{*}{Mask Location} & \multicolumn{2}{c}{Baseline} & \multicolumn{2}{c}{RNP} & $Neurons \downarrow$ \\ \cline{2-6} 
 & ASR (\%) & CA (\%) & ASR (\%) & CA (\%) & - \\ \hline
Block 1 & 100 & 93.40 & 9.98 & 73.37 & 56 \\ 
Block 2 & 100 & 93.40 & 9.40 & 88.60 & 67 \\ 
Block 3 & 100 & 93.40 & 9.01 & 93.10 & 62 \\ 
Block 4 & 100 & 93.40 & 6.50 & 93.61 & 55 \\ 
All Block & 100 & 93.40 & \textbf{0.20} & \textbf{92.22} & \textbf{41} \\ \bottomrule
\end{tabular}
\label{tab:filter_location}
\end{table}

\section{Exploration with the Unlearned Model} \label{ap:d}
Here, we will show that the unlearned model (NU-model) produced by our NU technique can be leveraged not only to improve existing defense methods but also to assist with trigger recovery, backdoor label detection, and backdoor sample detection.

\subsection{Improving Backdoor Removal} \label{ap:NU_removal}
Previous work proposed fine-tuning on the recovered trigger (denoted as 'NC+FT') \cite{wang2019neural}, trigger unlearning (denoted as 'NC+ABL') \cite{li2021anti}, and fine-pruning (denoted as 'FP') \cite{liu2018fine} to repair the backdoored model. To further explore our NU technique, we propose to perform backdoor removal on the NU-unlearned model. We set NC+FT as the baseline. The experiment results in Table \ref{tab:NU_results} show that using the unlearned model can significantly improve the performance of the three defenses against all attacks. More specifically, in comparison to the baseline (i.e., NC+FT), adding NU can further lower the average ASR for NC+FT (row 2), NC+ABL (row 3), and FP (row 4) by 4.18\%, 3.78\%, and 6.01\%, respectively. Meanwhile, NU also boosts clean accuracy to a certain extent. These results verify that NU can indeed be a central part of effective backdoor removal.

\begin{table*}[!htbp] 
\small
\centering
\caption{Performance of 3 defense methods with/without our proposed NU respectively against backdoor attacks on CIFAR-10. ASR: attack success rate (\%); CA: clean accuracy (\%); \textit{AvgDev}: the average \% deviation in ASR/CA compared to the baseline 'NC+FT'.}

\vskip 0.15in
\begin{adjustbox}{width=0.9\linewidth}
\begin{tabular}{c|c|cccccccccc|cc}
\toprule
Metric & Defense & BadNets & Trojan & Blend & CL & SIG & Dynamic & WaNet & FC & DFST & AWP & AvgDev \\ \hline
\multirow{3}{*}{ASR} & NC+FT (Baseline) & 8.31 & 1.64 & 3.71 & 3.51 & 5.62 & 9.71 & 8.19 & 43.69 & 12.50 & 3.20 & -\\ \cline{2-13} 
 & NU+NC+FT & 0.72 & 0.78 & 0.24 & 0.58 & 1.18 & 8.67 & 1.10 & 34.58 & 9.60 & 0.87 & $\downarrow$~\textbf{4.18} \\
 & NU+NC+ABL & 0.28 & 1.02 & 1.87 & 0.24 & 3.37 & 46.86 & 7.79 & 0 & 0.9 & 0.01 & $\downarrow$~\textbf{3.78}\\
 & NU+FP & 0.37 & 1.42 & 0 & 6.48 & 0.04 & 11.16 & 10.19 & 4.91 & 4.89 & 0.56 & $\downarrow$~\textbf{6.01} \\ \hline
\multirow{3}{*}{CA} & NC+FT (Baseline) & 93.21 & 92.96 & 92.63 & 92.28 & 82.06 & 91.28 & 83.32 & 87.62 & 87.93 & 94.16 &  - \\ \cline{2-13} 
 & NU+NC+FT & 92.13 & 93.14 & 92.64 & 83.79 & 80.36 & 91.08 & 88.32 & 92.76 & 88.93 & 94.70 & $\uparrow$~\textbf{0.03} \\
 & NU+NC+ABL & 93.92 & 93.28 & 91.32 & 93.23 & 88.60 & 81.19 & 93.30 & 85.06 & 93.22 & 94.36 & $\uparrow$~\textbf{1.00}\\
 & NU+FP & 89.94 & 85.30 & 92 & 92.57 & 93.88 & 86.84 & 92.23 & 91.87 & 92.58 & 94.53 & $\uparrow$~\textbf{1.42} \\ \bottomrule
\end{tabular}
\end{adjustbox}
\vskip -0.1in
\label{tab:NU_results}
\end{table*}

\subsection{Improving Trigger Recovery} \label{ap:NU_recover}
\begin{figure*}[!tp]
    \centering
    \begin{minipage}[t]{0.49\textwidth}
    \centering
    \includegraphics[width = .99\linewidth]{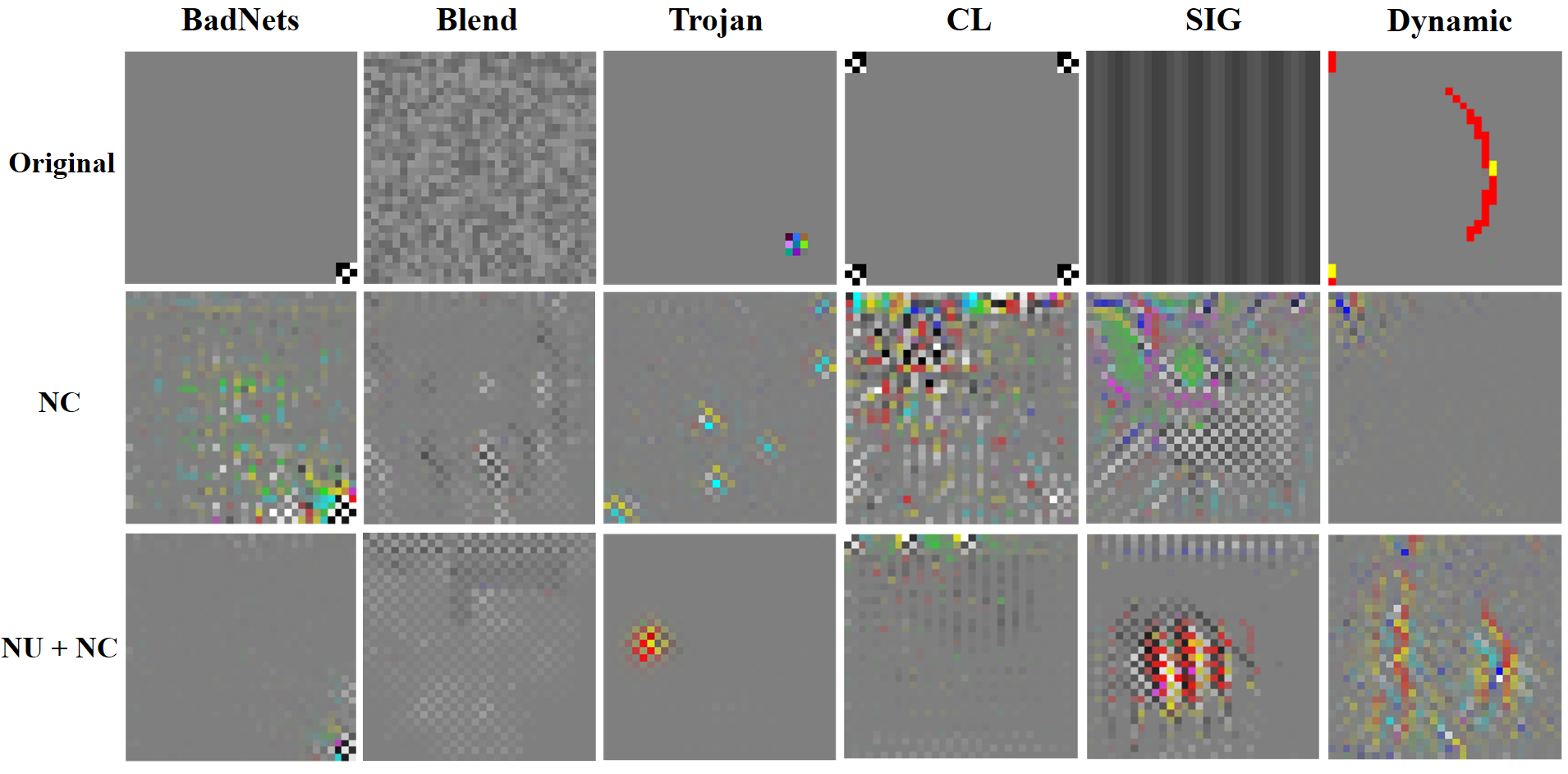}
    \caption{Side-by-side comparison of the original trigger patterns and their recovered versions by NC on the backdoored models and by our NU (NU+NC) on the unlearned models.}
    \label{fig:NU_recover}
    \end{minipage}
    \hfill
    \begin{minipage}[t]{0.49\textwidth}
    \centering
    \includegraphics[width = .79\linewidth]{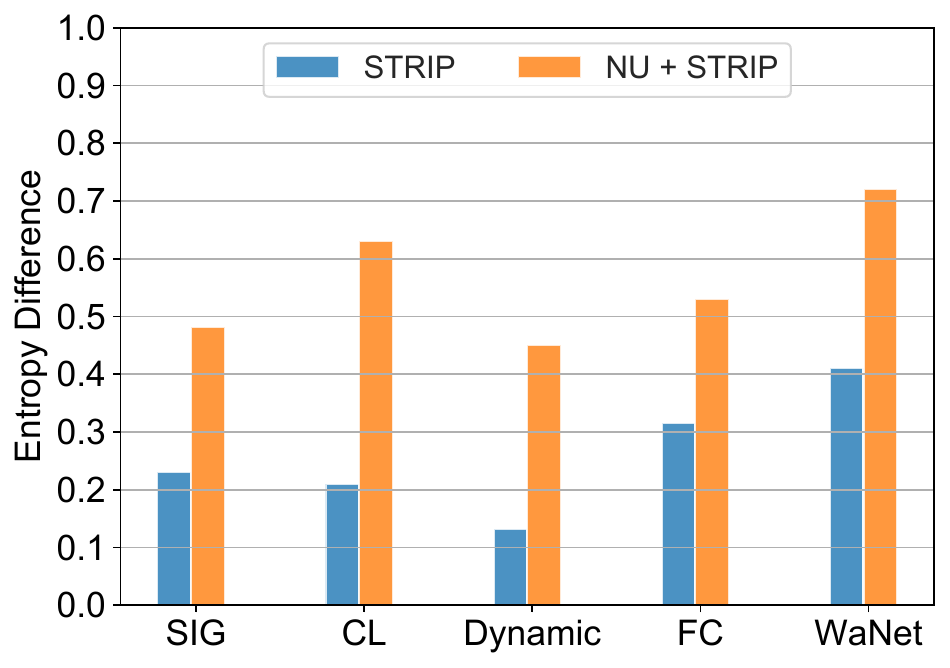}
    \caption{Overall entropy difference between clean and backdoor examples. A greater difference in entropy indicates a better detection performance.}
     \label{fig:NU_strip}
    \end{minipage}
\end{figure*}

Figure \ref{fig:NU_recover} shows a few examples of recovered triggers. We observe that, compared to the NC alone (second row), the triggers recovered by the NU-unlearned model present a more precise trigger quality and a reasonable size for almost all attacks. NC alone fails to recover the trigger on CL, SIG, and Dynamic, due to the recovered trigger having more image noise. We speculate that the success of 'NU+NC' is because the unlearned model contains fewer clean but more backdoor neurons than the original backdoored model. In other words, the unlearned model makes trigger reverse engineering easier.

\subsection{Improving Backdoor Label Detection} \label{ap:NU_label}
Table \ref{tab:NU_label} shows the NU-unlearned model can efficiently boost the performance of backdoor label detection for Neural Cleanse (NC). Interestingly, NU alone achieves the best detection rate by 100\% on all the attacks. This is because, as clean neurons are gradually removed from the model, the backdoor class naturally arises. As a result, the samples are predicted by the unlearned model to be of the backdoor class.

\begin{table*}[!tp]
\small
\centering
\caption{Detection accuracy (top-1, \%) of the backdoor label. }
\begin{adjustbox}{width=0.8\linewidth}
    \begin{tabular}{c|ccccccccccc}
    \toprule
    Types & BadNets & Trojan & Blend & CL & SIG & Dynamic & WaNet & FC & DFST & AWP & \\ \hline
    NC & 65 & 95 & 90 & 90 & 15 & 80 & 85 & 0 & 20 & 95 & \\
    NU+NC & 100 & 100 & 100 & 100 & 100 & 100 & 100 & 90 & 95 & 100 \\
    NU  & 100 & 100 & 100 & 100 & 100 & 100 & 100 & 100 & 100 & 100 \\ 
    \bottomrule
    \end{tabular} 
\end{adjustbox}
\vskip -0.1in
 \label{tab:NU_label}
\end{table*}

\subsection{Improving Backdoor Sample Detection} \label{ap:NU_strip}
Figure \ref{fig:NU_strip} shows the average relative entropy between the outputs of the clean and the backdoored examples by the original STRIP and our NU combined with STRIP (denoted as NU+STRIP). We find that NU+STRIP creates a greater relative entropy (nearly 0.2) than the ordinary STRIP. The greater the relative entropy, the better the detection performance on filtering poisoned examples. Our NU amplifies the gap between clean and backdoor outputs in relative entropy, consequently contributing to detecting more advanced triggers.

\end{document}